%% file: main.tex
\documentclass{article}

 \usepackage[preprint]{neurips_2025}
 \usepackage{graphicx}
 \usepackage{amsmath}
 \usepackage{dsfont}
 \usepackage{amsthm}
 \usepackage{amssymb}
\theoremstyle{plain}
\newtheorem{theorem}{Theorem}

\usepackage[utf8]{inputenc} 
\usepackage[T1]{fontenc}    
\usepackage{hyperref}       
\usepackage{url}            
\usepackage{booktabs}       
\usepackage{amsfonts}       
\usepackage{nicefrac}       
\usepackage{microtype}      
\usepackage{xcolor}         
\usepackage{cleveref}
\crefname{appendix}{App.}{Apps.}
\crefname{equation}{Eq.}{Eqs.}
\crefname{figure}{Fig.}{Figs.}
\crefname{table}{Tab.}{Tabs.}
\crefname{section}{Sec.}{Secs.}
\usepackage{xcolor}

\title{A Quantitative Study of Sustained Focus in Large Language Models via Repetitive Deterministic Prediction Tasks}

\author{
  Wanda Hou \\
  EdenCode Inc. \\
  \texttt{hwanda@edencode.ai} \\
  \And
  Leon Zhou \\
  Stevenson School \\
  \texttt{lzhou26@stevensonschool.org} \\
  \\
  \And
  Hong-Ye Hu \\
  Harvard University \\
  \texttt{hongyehu@fas.harvard.edu} \\
  \And
  Yubei Chen\\
  UC Davis \\
  \texttt{ybchen@ucdavis.edu} \\
  \And
  Yi-Zhuang You \\
  EdenCode Inc. \\
  \texttt{yzyou@edencode.ai} \\
  \And
  Xiao-Liang Qi \\
 Path Integral Technology, Inc. \\
  \texttt{phynics@pathintegral.xyz} \\
}

\begin{document}

\maketitle

\begin{abstract}
We investigate the performance of large language models (LLMs) on repetitive deterministic prediction tasks and study how the sequence accuracy rate (SAR) scales with output length. Each such task involves the repetition of the same operation $N$ times. Examples of such tasks include letter replacement in letter strings following a given rule, integer addition, and multiplication of string operators in many-body quantum mechanics. If the LLM performs the task by a simple repetition algorithm, the success rate would follow an exponential decay with sequence length. In contrast, our experiments on leading LLMs reveal a crossover that is sharper than exponential: $-\log\mathrm{SAR}$ grows super-linearly with $N$, and accuracy collapses around a characteristic length $N_*$, the accuracy cliff that separates reliable from unreliable generation. The hypothesis of independent per-step errors is rejected for every model and task we studied. The crossover is well described by a double-exponential accumulation law, $\mathrm{SAR}=\exp(-\beta_0 N\alpha^{N-1})$, whose crossover scale $N_*$ does not depend on the functional form chosen to fit it. To interpret this behaviour we introduce a minimal effective model in which step-correctness variables interact through dense random couplings and compete with an external field set by the prompt. Solved by direct enumeration, the model reproduces the super-linear error accumulation and the accuracy cliff qualitatively, and it assigns to each model--task pair two interpretable parameters, an intrinsic error rate and an error-accumulation factor.

\end{abstract}

\section{Introduction}

\emph{Accuracy rate} is among the most fundamental and widely adopted evaluation metrics in machine learning, especially for classification and other discriminative tasks. Its appeal lies in the clarity of its definition: each input has a well-defined ground-truth label, and the model's prediction can be judged unambiguously as correct or incorrect. This binary criterion makes accuracy a natural measure of performance, enabling direct comparison across models and datasets.

However, this paradigm does not directly extend to large language models (LLMs). Many LLM applications, such as machine translation, summarization, or dialogue, are inherently sequence-to-sequence generation problems. In these settings, the model generates text probabilistically, token by token, sampling from a distribution of plausible continuations. Crucially, there is rarely a single ``correct'' output: multiple valid translations, summaries, or conversational responses may exist. Consequently, simple exact-match accuracy is insufficient for evaluation, and researchers rely instead on a suite of task-specific metrics (e.g., BLEU, ROUGE, perplexity, or human judgment)~\citep{papineni-etal-2002-bleu,lin-2004-rouge,liang2022holistic} that capture fluency, coherence, and relevance.

Yet this limitation is context-dependent. In scientific domains---including mathematics, physics, and computer science---the nature of the task often changes. Here, many problems admit a unique correct solution: a definite numerical value, a closed-form expression, or a precise code output. For instance, the question ``what is $2342342 + 3442342$'' has only one acceptable answer, and the model's response must exactly match the correct sequence of digits to be considered correct~\citep{cobbe2021training,hendrycks2021measuring,lewkowycz2022solving}. In such applications, the probabilistic flexibility of LLMs does not diminish the importance of exact correctness. The evaluation once again reduces to a binary criterion: either the model outputs the correct sequence or it does not.

\begin{figure}
    \centering
    \includegraphics[width=0.6\linewidth]{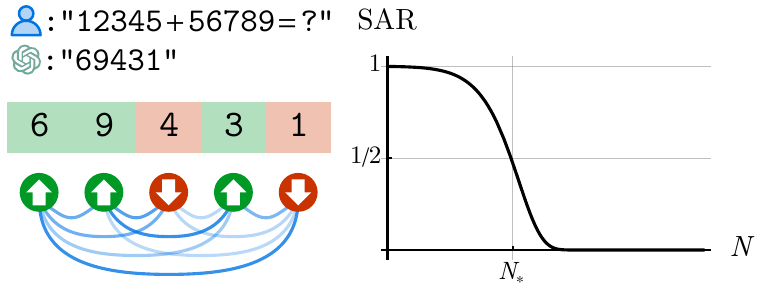}
    \caption{Illustration of the experimental setup and theoretical framework.
The left panel shows examples of deterministic sequence prediction tasks (e.g., arithmetic problems) used to evaluate large language models (LLMs), where each token’s correctness is represented by a binary variable (Ising spin). Errors at different positions are modeled as Ising spins interacting through dense random couplings, an effective description of how correlations and noise propagate during sequence generation. The right panel displays the typical behavior of the Sequence Accuracy Rate (SAR) (i.e.~the probability for the whole output sequence to be correct) as a function of the sequence length $N$: SAR remains high for short sequences, then drops beyond a characteristic crossover scale $N_*$, forming the accuracy cliff.}
    \label{fig:illustration}
\end{figure}

This observation motivates the introduction of a new evaluation concept tailored for LLMs in scientific applications: the \emph{Sequence Accuracy Rate (SAR)}.
SAR measures the proportion of problem instances for which the LLM produces the exact correct sequence as output.
By aligning the evaluation of LLMs with the well-established standard of accuracy in classification, SAR provides a rigorous and interpretable metric for assessing LLMs in domains where precision and correctness are paramount, as illustrated in \cref{fig:illustration}.

LLMs have achieved remarkable progress across a wide range of tasks, yet their reliability on structured, multi-step reasoning remains poorly understood~\citep{wei2022chain,havrilla2024glore}. While in generative modeling the challenge has often been framed as a tradeoff between diversity and realism~\citep{astolfi2024consistency}, multi-step reasoning tasks demand consistent correctness across sequential subtasks. In the simplest case, success in these settings is approximately \emph{factorizable}: for tasks whose output positions are computed independently (such as cyclic letter replacement), the probability of solving the overall problem is the product of per-step accuracies, making performance highly sensitive to individual step error rates. Other tasks introduce explicit dependencies between steps---integer addition couples digits through long-range carry propagation, and Pauli string multiplication couples sites through a shared global phase---so that the strict factorization holds only as a leading-order approximation. This range of dependency structures is exactly what allows us to probe whether models exploit compositional structure and execute each step independently, rather than treating complex problems monolithically~\citep{lake2018generalization}.

To probe this challenge, we introduce a minimal and fully deterministic evaluation class where each problem instance admits a unique ground-truth solution and complexity can be tuned by a single parameter $N$. We present three complementary benchmarks: (i) \emph{cyclic letter replacement task}, a purely symbolic problem over a finite alphabet that isolates stepwise error propagation without memory; (ii) \emph{integer addition}, which tests the handling of local carry rules and long-range carry propagation across digits; and (iii) \emph{Pauli string multiplication}, which requires tracking both local symbol rules and a global phase factor, thereby testing the model’s ability to maintain a lightweight memory across many steps All three benchmarks share the key property that overall complexity decays multiplicatively with sequence length $N$, providing a clean probe of cumulative error accumulation in deterministic reasoning~\citep{wang2022self,liu2023lost,turpin2023language,wan2025fano}.

Beyond empirical evaluation, we propose a statistical-physics interpretation of the observed behavior of SAR, drawing an analogy to an Ising model in which the correctness variables of the individual task positions influence each other through effective random couplings. Solved by direct enumeration, this minimal model reproduces the super-linear error accumulation and the accuracy cliff qualitatively and provides a two-parameter description of each model--task pair. The model is an \emph{effective}, phenomenological description and should not be read as a microscopic theory of the network; what the data establish is the super-linear accumulation of errors, and the mechanism behind it remains open. Together, our benchmarks and theoretical framework establish a controlled setting for studying how LLMs manage compositional reasoning, memory, and robustness in deterministic problem families.

\section{Experiment}\label{sec:experiment}

We prompted a set of large language models (LLMs)---\texttt{gpt-5}, \texttt{gemini-2.5-pro}, \texttt{gemini-2.5-flash}, \texttt{grok-4}, and \texttt{claude-4-sonnet}---to perform a series of deterministic sequence prediction tasks that each admit a unique correct solution.
These tasks provide a well-defined setting for evaluating the sequence accuracy rate (SAR), enabling a quantitative analysis of how the accuracy of generated sequences scales with the problem size, measured by the output sequence length. During evaluation, all models were tested in a closed setting with external tools such as code execution and web access explicitly disabled. Concretely, every query was issued through a plain text completion/chat call with no tool-use, function-calling, or code-interpreter interfaces enabled (see the source code); consequently the models cannot invoke external calculators or solvers, and all results reflect the models' internal, in-context reasoning capabilities only.

\paragraph{Estimation of SAR.} For each task and each sequence length $N$ we pose $n$ independent problem instances in batches of 10 freshly sampled problems per prompt, typically 10 batches per length ($n=100$; a subset of \texttt{grok-4} runs use fewer, $n\le30$). Throughout the paper the empirical SAR at length $N$ is the \emph{fraction of instances whose complete answer is correct}, $\widehat{\mathrm{SAR}}(N)=k/n$; the same estimator is used for every task and every figure. A batch whose reply could not be parsed into answers at all (an empty or malformed response) is treated as a protocol failure and excluded from $n$. Such batches make up between $0$ and $9\%$ of the total depending on the model and occur mostly at large $N$; counting their instances as wrong instead shifts the fitted crossover scales by at most $5\%$. Because the ten instances of a batch are generated in one reply, their outcomes are positively correlated. The between-batch variance of the batch accuracy exceeds the binomial value by a design effect $\hat c$, which we estimate for every model--task pair: $\hat c=1.0$--$1.9$ for letter replacement and addition, $2.1$--$4.2$ for the reasoning models, and $4.8$--$5.9$ for Pauli multiplication. All error bars are therefore $95\%$ Wilson intervals computed with the effective sample size $n/\hat c$, all fits maximise the binomial likelihood of the instance counts, and the information criteria are rescaled accordingly (quasi-AIC, $\mathrm{QAIC}=2P-2\log L/\hat c$ with $P$ the number of fitted parameters; see \cref{app:comparison}). Sampling temperature was set to $0$ wherever the API exposes it; the reasoning models (\texttt{gpt-5}, \texttt{grok-4}) use their fixed internal sampling and do not expose a temperature parameter. While the quantitative results may vary slightly with prompt phrasing or instruction style, the qualitative behaviors reported here are robust across formulations. The prompt templates, per-model settings, raw records and the complete analysis pipeline are available at
\href{https://github.com/EdenCodeInc/PyCliffordMCP/tree/main/dev/pauli_string_multiplication}{\texttt{GitHub}}.

\subsection{Cyclic Letter Replacement}

We start with a simple \emph{cyclic letter replacement} task defined over an alphabet
\begin{equation}
\mathcal{A} = \{\mathrm{A}, \mathrm{B}, \mathrm{C}, \ldots\}.
\end{equation}
Each letter is mapped to an integer index $x \in \{0,1,\ldots,|\mathcal{A}|-1\}$.
The transformation rule is a cyclic shift by one modulo the alphabet size:
\begin{equation}
y = (x+1) \bmod |\mathcal{A}|.
\end{equation}
The output is then mapped back to a letter in $\mathcal{A}$. Each position in the input string is transformed independently, so for a string of length $N$ the task is simply to apply the shift rule to each site.

Formally, given an input string
\begin{equation}
S = \big(x_1, x_2, \ldots, x_N\big), \quad x_j \in \{0,\ldots,|\mathcal{A}|-1\},
\end{equation}
the transformed output string is
\begin{equation}
T = \big((x_1+1)\bmod |\mathcal{A}|,\ (x_2+1)\bmod |\mathcal{A}|,\ \ldots,\ (x_N+1)\bmod |\mathcal{A}|\big).
\end{equation}

\begin{figure}[ht]
    \centering
    \includegraphics[width=0.65\linewidth]{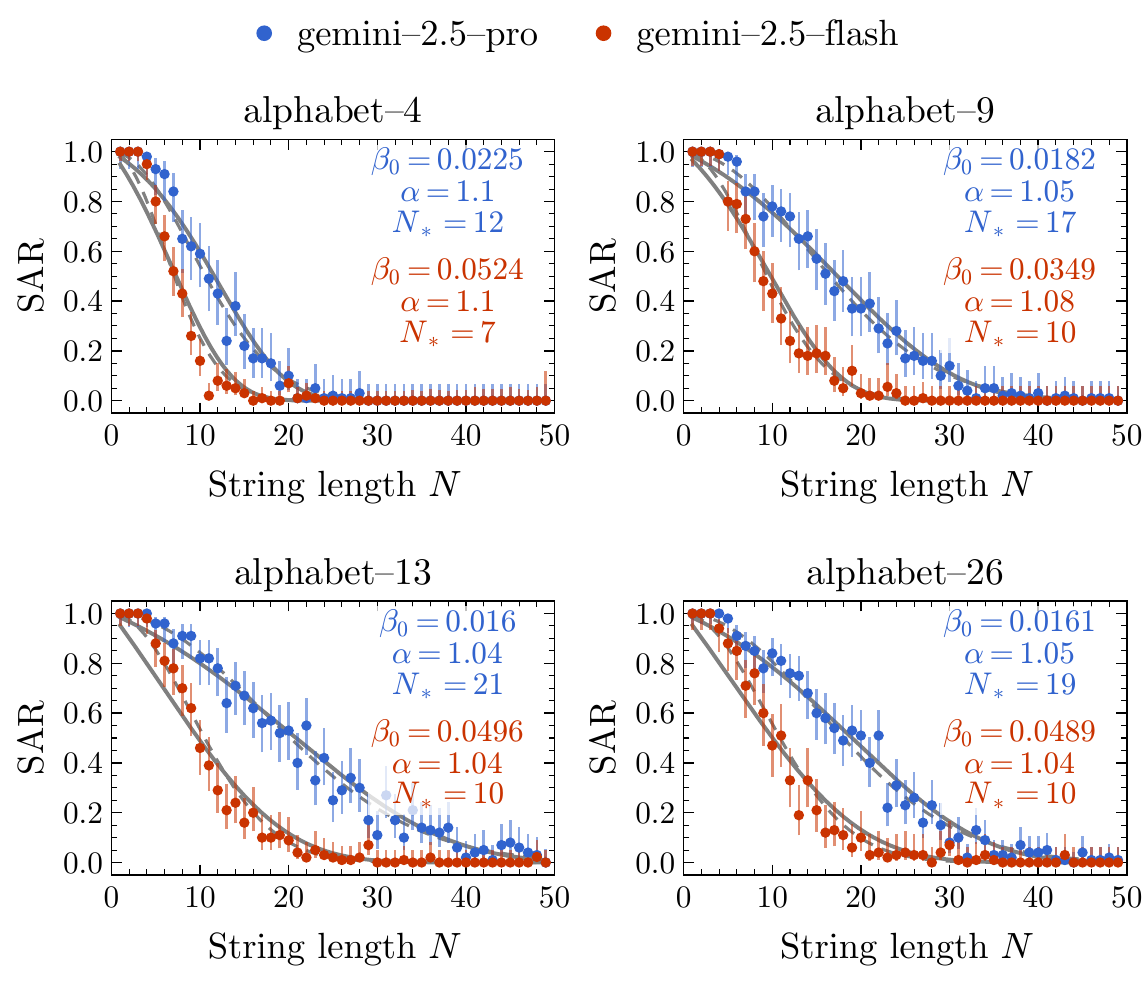}
    \caption{Cyclic letter replacement benchmark on \texttt{gemini-2.5-pro} and \texttt{gemini-2.5-flash} with alphabet size $|\mathcal{A}|=4,9,13$ and $26$. Points: fraction of correct instances (up to $n=100$ per length); error bars: $95\%$ Wilson intervals with the batch design effect. Solid curves: maximum-likelihood fit of the accumulation law \cref{emp_sar}, with the fitted $\beta_0$, $\alpha$ and crossover scale $N_*$; dashed: the quadratic law $e^{-aN-bN^2}$ (\cref{sec:comparison}).}
    \label{fig:data-alph}
\end{figure}

For example, for $\mathcal{A}={A,B,C,D}$:
\begin{center}
\begin{tabular}{lcl}
Input: & ADBAA & $\Rightarrow$ Output: BACBB \\
Input: & DABDC    & $\Rightarrow$ Output: ABCAD
\end{tabular}
\end{center}

\begin{figure}[ht]
    \centering
    \includegraphics[width=0.95\linewidth]{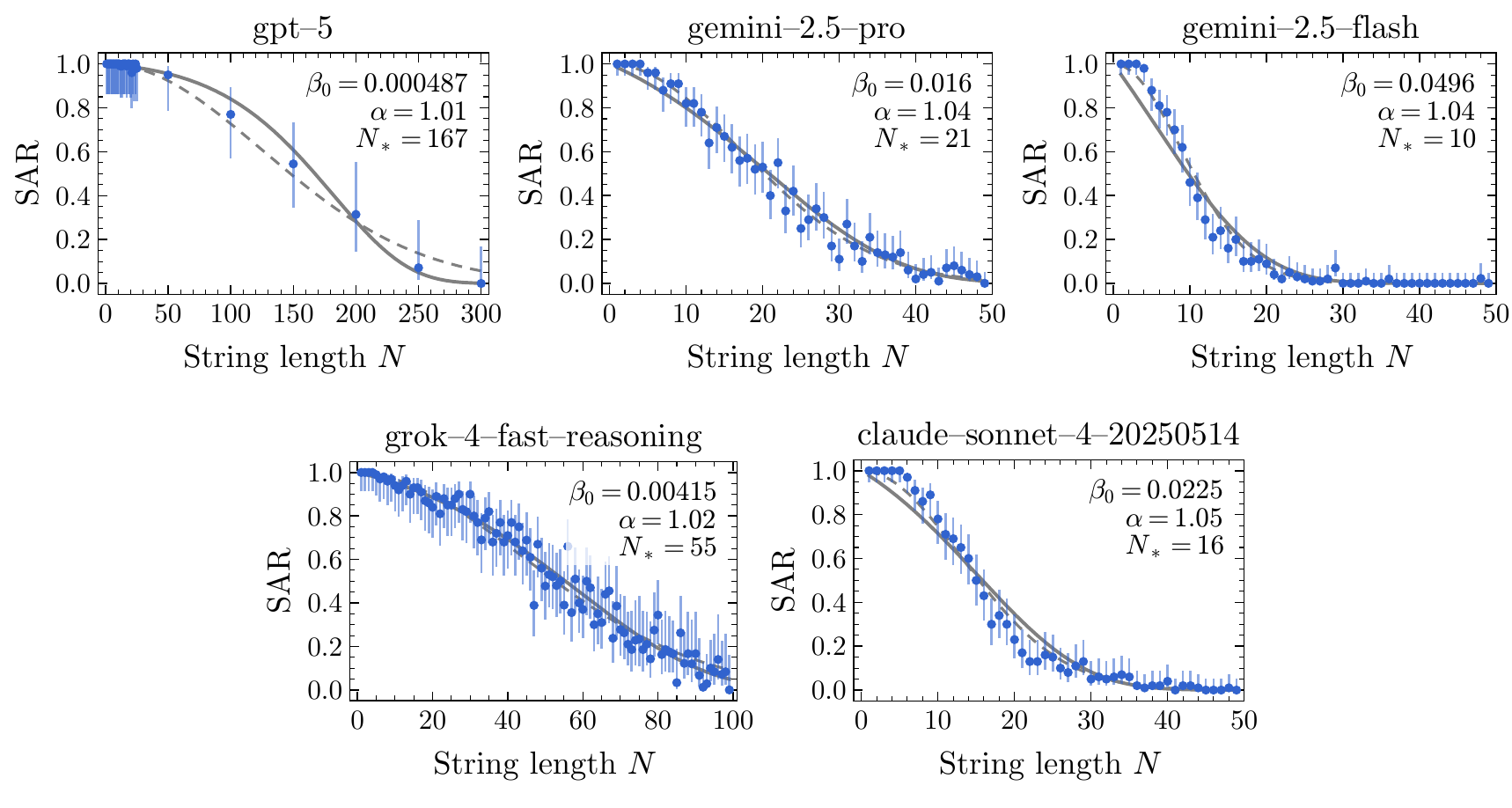}
    \caption{Cyclic letter replacement with alphabet size $|\mathcal{A}|=13$ across different models. Points: fraction of correct instances (up to $n=100$ per length; $n\le30$ for some \texttt{grok-4} lengths); error bars: $95\%$ Wilson intervals with the batch design effect. Solid: fit of \cref{emp_sar}; dashed: quadratic law.}
    \label{fig:data-alph-13}
\end{figure}

For this task, the results are presented in \cref{fig:data-alph} and \cref{fig:data-alph-13}.
\cref{fig:data-alph} shows results from \texttt{gemini-2.5-pro} and \texttt{gemini-2.5-flash} with alphabet sizes $|\mathcal{A}|=4,9,13,26$.
\cref{fig:data-alph-13} compares \texttt{gpt-5}, \texttt{gemini-2.5-pro}, \texttt{gemini-2.5-flash}, \texttt{grok-4}, and \texttt{claude-4-sonnet} at $|\mathcal{A}|=13$.

\subsection{Integer Addition}

Next, we consider \emph{integer addition} with a fixed number of digits.
The task probes arithmetic consistency and multi-digit carry propagation—requiring sequential reasoning rather than purely single-site transformations.

Each query presents an addition problem of the form
\begin{equation}
A + B,
\end{equation}
where \(A,B \in \mathbb{N}\) are positive integers with the same digit lengths.
The model is asked to compute the sum and return the result as a plain numeric string.

For example at $N=4$:

\begin{center}
\begin{tabular}{lcl}
Input: & \(\texttt{1234} + \texttt{5678}\) & $\Rightarrow$ Output: \texttt{6912} \\
Input: & \(\texttt{9999} + \texttt{0001}\)    & $\Rightarrow$ Output: \texttt{10000}
\end{tabular}
\end{center}

\begin{figure}[ht]
    \centering
    \includegraphics[width=0.65\linewidth]{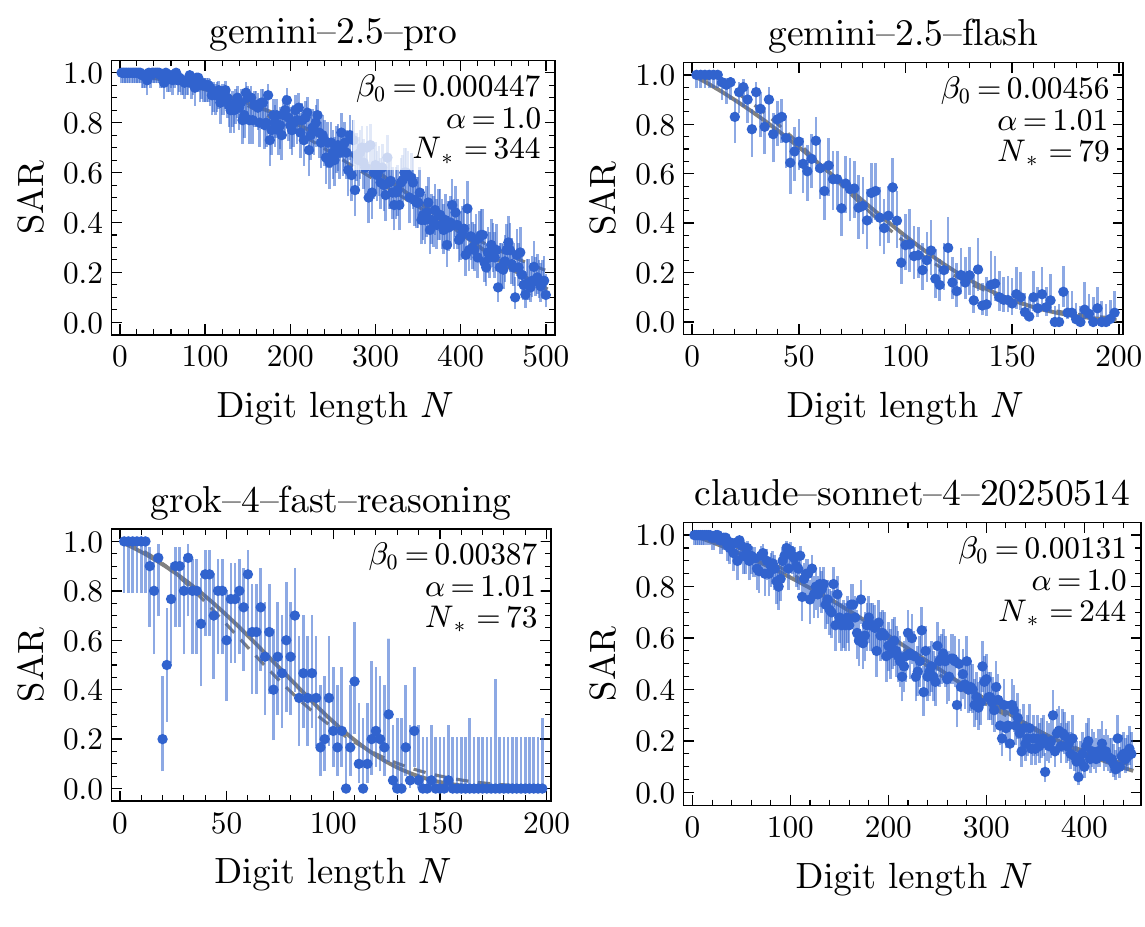}
    \caption{Integer addition benchmark across different models. Points: fraction of correct instances (up to $n=100$ per length; $n\le30$ for \texttt{grok-4}); error bars: $95\%$ Wilson intervals with the batch design effect. Solid: fit of \cref{emp_sar}; dashed: quadratic law.}
    \label{fig:data-add}
\end{figure}

The corresponding results are shown in \cref{fig:data-add},
evaluating \texttt{gemini-2.5-flash}, \texttt{gemini-2.5-pro}, \texttt{grok-4}, and \texttt{claude-4-sonnet}.

\subsection{Pauli Algebra}

As a third deterministic benchmark, we consider \emph{Pauli string multiplication} task with tunable string length. The Pauli operators $X$, $Y$, and $Z$ are $2\times 2$ matrices
\begin{align}
    X=\left(\begin{array}{cc}0&1\\1&0\end{array}\right),~Y=\left(\begin{array}{cc}0&-i\\i&0\end{array}\right),~Z=\left(\begin{array}{cc}1&0\\0&-1\end{array}\right),
\end{align}
They originate as generators of the Lie algebra $\mathfrak{su}(2)$, up to constants. For our purposes we treat them algebraically, using only their multiplication rules and phase factors. The identity operator is denoted $I$. The single-qubit multiplication rules are
\begin{equation}
X \times X = Y \times Y = Z \times Z = I,
\end{equation}
\begin{equation}
X \times Y = iZ, \quad Y \times X = -iZ,
\end{equation}
\begin{equation}
Y \times Z = iX, \quad Z \times Y = -iX,
\end{equation}
\begin{equation}
Z \times X = iY, \quad X \times Z = -iY,
\end{equation}

together with $I \times P = P$ for $P \in \{I,X,Y,Z\}$.

A Pauli string on $N$ sites is
\begin{equation}
S = \alpha \bigotimes_{\ell=1}^{N} P_\ell,
\end{equation}
where $\alpha \in \{\pm 1,\pm i\}$ is a global phase and $P_\ell \in \{I,X,Y,Z\}$.
Because Pauli operators acting on different sites commute, the product of two strings reduces to independent local multiplications together with accumulation of the global phase:
\begin{equation}
\left(\alpha_1 \bigotimes_{\ell=1}^{N} P_\ell \right)
\left(\alpha_2 \bigotimes_{\ell=1}^{N} Q_\ell \right)
= \Big(\alpha_1 \alpha_2 \prod_{\ell=1}^{N} \phi_\ell \Big)
\bigotimes_{\ell=1}^{N} R_\ell,
\end{equation}
where $P_\ell \times Q_\ell = \phi_\ell R_\ell$ with $\phi_\ell \in \{\pm 1,\pm i\}$ and $R_\ell \in \{I,X,Y,Z\}$.

This structure makes the task explicitly factorizable:
each site requires a local Pauli multiplication and a phase update,
while the overall result depends on consistent accumulation of all local outcomes. See \cref{app:pauli_sar} for a theoretical complexity analysis.

The need to consistently maintain and update the global phase highlights the role of \emph{memory handling} in multi-step reasoning. Unlike purely sitewise problems, errors here can accumulate not only locally but also through the global phase register.

\begin{figure}[t]
    \centering
    \includegraphics[width=0.65\linewidth]{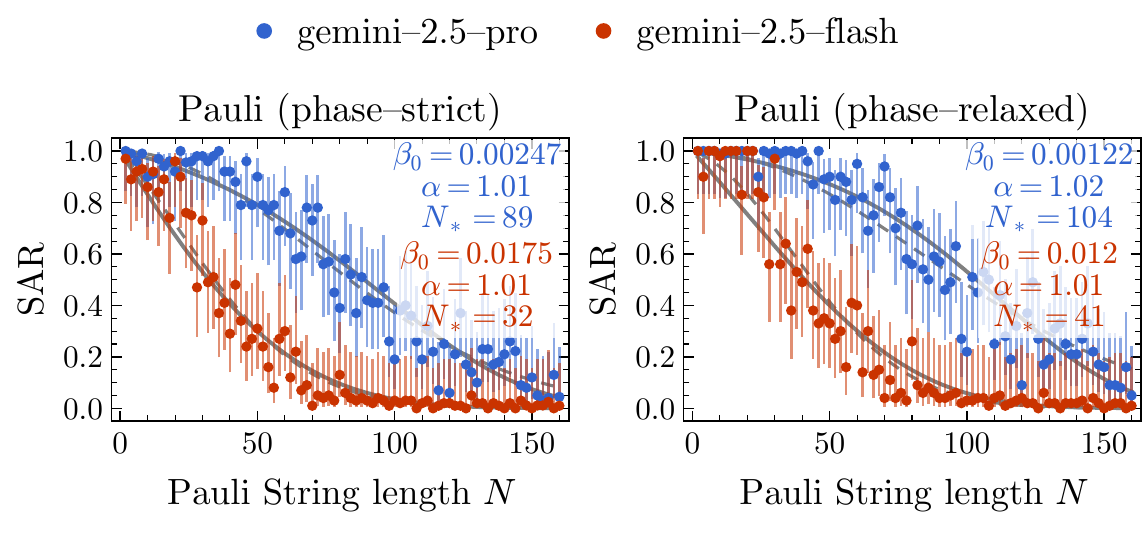}
    \caption{Pauli string multiplication benchmark on \texttt{gemini-2.5-pro} and \texttt{gemini-2.5-flash} evaluated under strict (phase-matched) and relaxed (phase-ignored) criteria. Points: fraction of correct instances (up to $n=100$ per length); error bars: $95\%$ Wilson intervals with the batch design effect ($\hat c\approx5$ for this task). Solid: fit of \cref{emp_sar}; dashed: quadratic law.}
    \label{fig:data-pauli}
\end{figure}

For this task, the results are presented in \cref{fig:data-pauli},
evaluating \texttt{gemini-2.5-pro} and \texttt{gemini-2.5-flash} under two evaluation criteria:
one requiring both the operator sequence and accumulated phase to be correct,
and a relaxed criterion considering only the operator sequence while ignoring phase errors.

\section{Analysis}

\subsection{Empirical Formula}

In our experiments, we observed that the Sequence Accuracy Rate (SAR) depends strongly on the output sequence length $N$. Assuming independent error events at the individual steps of the task, the baseline expectation is that the sequence accuracy rate decays exponentially with the sequence length $N$:
\begin{equation}\label{exp_sar}
    \text{SAR}(N) = \mathrm{e}^{-\beta N},
\end{equation}
where $\beta>0$ denotes a constant per-step error rate.
However, as we have seen in the above experiments, the empirical results deviate significantly from this simple picture.
Instead of a smooth exponential decay, SAR exhibits a crossover:
it remains nearly perfect for short sequences and then drops to zero around a characteristic length scale $N_{*}$, more rapidly than an exponential would.

This observation indicates that the errors at different steps are not independent: earlier steps affect the error rate of later ones. To capture this history dependence, we treat the per-step error rate as a function of the sequence length $N$ rather than a constant and adopt the following empirical fit:
\begin{equation}\label{emp_sar}
    \text{SAR}(N) = \mathrm{e}^{-\beta(N)N},
    \quad \beta(N) = \beta_0\, \alpha^{N-1},
\end{equation}
where $\beta_0>0$ quantifies the \emph{intrinsic} error rate,
and $\alpha$ represents the \emph{error accumulation factor},
which characterizes how local errors compound as the sequence length increases.
When $\alpha = 1$, the model reduces to the standard exponential decay expected from independent errors. For $\alpha > 1$, however, the effective error rate $\beta(N)$ grows exponentially with $N$ even when $\beta_0$ is small, leading to a rapid drop in accuracy once $\beta(N)$ becomes of order one. The onset of this accuracy cliff occurs at a characteristic sequence length
\begin{equation}\label{eq:Nc}
    N_*\simeq 1+\frac{\log(1/\beta_0)}{\log\alpha},
\end{equation}
beyond which scale the model's reliability collapses.

\paragraph{Is the functional form warranted?}\label{sec:comparison}
Two questions must be kept apart: whether the data reject the independent-error law \cref{exp_sar}, and whether they single out the particular $N$-dependence of \cref{emp_sar}. To answer both we fitted, by binomial maximum likelihood, \cref{emp_sar} together with its one-parameter special case $\alpha=1$ and with three alternative two-parameter crossover laws: the quadratic exponent $\exp(-aN-bN^2)$, which is the leading-order prediction of the model of \cref{sec:model} and the natural competitor suggested by the perturbative expansion; a stretched exponential $\exp[-(N/N_0)^\gamma]$; and a logistic crossover $[1+e^{(N-N_0)/w}]^{-1}$. The laws are compared by quasi-AIC and quasi-BIC (rescaled by the design effect $\hat c$) and by leave-one-length-out predictive likelihood; the full table is given in \cref{app:comparison}. On the first question the answer is clear. The exponential law is rejected on every one of the 19 model--task curves, by $\Delta\mathrm{QAIC}\ge 35$ and up to $1.6\times10^{3}$, so the super-linear growth of $-\log\mathrm{SAR}$ with $N$ is an established feature of the data. On the second question the answer is negative. The four crossover laws describe the data comparably: the stretched exponential has the lowest QAIC on 14 of the 19 curves, \cref{emp_sar} on 3 and the logistic on 2, and the differences, $|\Delta\mathrm{QAIC}|\le1.6\times10^{2}$, are one to two orders of magnitude smaller than the margin against the exponential. The data alone therefore do not discriminate between crossover forms of comparable flexibility. We adopt \cref{emp_sar} as the working parametrisation for two reasons. It is the exponentiation of the leading-order correction of the model in \cref{sec:model}, and the quantity we use to compare models and tasks does not depend on the choice: the crossover scales $N_*$ obtained from the four laws agree to within $6\%$ (median over curves) and $15\%$ at most. The fitted accumulation factors are modest, $\alpha=1.003$--$1.10$, i.e.\ $\log\alpha$ between $3\times10^{-3}$ and $0.1$; the crossover is nevertheless far from exponential because $\alpha^{N-1}$ is evaluated at $N\sim N_*$, which ranges from $7$ to $344$ across our curves.

\subsection{Hypothesis and Statistical Model}\label{sec:model}

Empirically, the observed crossover length $N_*$ is typically on the order of $10\sim 300$---well within the attention window ($\sim 10^5$ or larger) of modern LLMs. This rules out limited context length as the primary cause of the accuracy cliff. We therefore propose an \emph{effective} model in which the super-linear accumulation of errors is captured by dense couplings among position-correctness variables. This choice is motivated by---though not derived from---the all-to-all structure of the self-attention mechanism. While the attention weights in next-token prediction are structured and typically sparse rather than uniformly distributed over all past tokens, the architecture nonetheless allows fluctuations and errors to spread through the network in an effectively all-to-all fashion, propagating globally across the entire generated sequence. As a result, small local errors can be amplified and spread through these dense interactions, effectively inducing collective fluctuations among the accuracies at different positions.

To analyze this mechanism, we abstract away the content of the output and focus solely on whether each position of the task is answered correctly. We introduce binary \emph{task-position correctness variables} $s_i = \pm 1$, $i=1,\dots,N$, with $s_i = +1$ if the $i$-th position of the task (a letter, a digit or a Pauli site, as opposed to a tokenizer token) is correct and $s_i = -1$ otherwise. We model the joint distribution of position errors $s = \{s_i\}$ using an effective Ising-like energy function,
\begin{equation}
E_J[s] = -\sum_{i<j} J_{ij}\, s_i s_j - h \sum_i s_i,
\end{equation}
where $h$ denotes an external field reflecting the baseline bias from the prompt toward correct generation, and $J_{ij}$ represent effective couplings between position-correctness variables. In the independent-position limit, $J_{ij}=0$ and the model predicts the exponential decay of Eq.~\eqref{exp_sar}. The rejection of that law in \cref{sec:comparison} implies correlations among the outcomes at different positions.

We model these correlations phenomenologically as random couplings drawn independently from a Gaussian ensemble, a form inspired by (but not derived from) the all-to-all structure of attention,
\begin{equation}
    \mathbb{E}_J[J_{ij}] = 0, \qquad
    \mathbb{E}_J[J_{ij}^2] = J_0^2,
\end{equation}
where $J_0$ characterizes the strength of internal interference or ``noise coupling.'' The random couplings capture, in an effective way, both constructive and destructive interference among position-correctness variables. Together with the external field $h$, which sets the intrinsic per-position accuracy, this formulation defines a statistical-physics model for the collective behavior of errors. Its energy function is that of the Sherrington--Kirkpatrick (SK) spin glass, with two important differences of interpretation that we make explicit.

\emph{Scaling of the couplings.} The coupling variance $J_0^2$ is independent of $N$. This differs from the conventional SK normalisation $J_0^2/N$, which is chosen to make the free energy extensive in the thermodynamic limit. No such limit is taken here, because $N$ is precisely the quantity varied in the experiments. The $N$-independent variance encodes the assumption that each additional output position couples to all earlier positions with a fixed characteristic strength, so that the total disorder energy grows as $N^{2}$ while the field energy grows as $N$. We stress that this growth is an \emph{assumption built into the effective model}, motivated by the all-to-all structure of attention but not derived from it, and that it should not be read as a prediction of spin-glass physics; with the standard $1/\sqrt N$ normalisation the same model produces no cliff (\cref{app:numerics}). In this sense the model is a dense random-coupling Ising model of SK type rather than the SK model proper, and we use the SK terminology only for the mathematical structure.

\emph{Symmetric couplings and position labels.} The couplings are symmetric, whereas autoregressive generation is directed: later outputs depend on earlier ones and not conversely. The energy model therefore describes the joint distribution of correctness outcomes and makes no statement about the generation dynamics; a causal variant with $J_{ij}\neq0$ only for $j<i$ is a natural extension that we leave to future work. Likewise, the index $i$ labels a task position and not a tokenizer token or an attention head, and nothing in the model identifies $J_{ij}$ with attention weights.

We emphasize that a genuinely disordered coupling is essential to the observed phenomenology, and that simpler alternatives are insufficient. A model containing only a field (uniform, or even a site-dependent random field $h_i$) but \emph{no} pairwise coupling factorizes over sites; its $\log\mathrm{SAR}$ is then strictly linear in $N$, reproducing the single exponential of \cref{exp_sar} with no crossover. Uniform \emph{ferromagnetic} couplings ($J_{ij}=J_0>0$) would instead drive a smooth mean-field transition, not the sharp, sample-to-sample crossover seen here. The random signs of $J_{ij}$ introduce \emph{frustration}---competing pairwise constraints that cannot all be satisfied simultaneously---while the field $h$ favors the all-correct configuration. It is the competition between this disorder (internal interference among generated positions) and the field (external conditioning from the prompt) that the model captures; the disorder contributes a super-linear-in-$N$ term to $-\log\mathrm{SAR}$, and hence an accuracy cliff. We note that the field explicitly breaks the $s\to-s$ symmetry, so there is no $\mathbb{Z}_2$ degeneracy between the all-correct and all-incorrect states: at small $N$ the field selects the all-correct configuration (paramagnetic regime), whereas at large $N$ the disorder energy ($\propto N^2$) overtakes the field, so the all-correct configuration is no longer favored---and, in the strong-disorder/weak-field regime, an incorrect configuration can become energetically competitive. This asymmetry is precisely what the field term is designed to encode.

For a given realization of the couplings $J_{ij}$, the probability of observing a configuration of position-correctness variables $s$ is
\begin{equation}
    p_J[s] \;=\; \frac{1}{Z_J} \, e^{-E_J[s]},
\end{equation}
with the partition function
\begin{equation}
    Z_J = \sum_{s} e^{-E_J[s]}.
\end{equation}
In particular, the SAR is associated with the probability of the fully correct configuration $s = (1,1,\ldots,1)$, averaged over disorder realizations of the couplings, which play the role of problem instances. The empirical estimator (fraction of correct instances) corresponds to the arithmetic disorder average $\mathbb{E}_J\,p_J[s=\mathbf 1]$; the replica calculation of \cref{app:sk-sar} is most naturally carried out for the geometric average,
\begin{equation}\label{ising_sar}
    \mathrm{SAR}_{\rm geo} \;:=\; \exp\!\Big( \mathbb{E}_J \big[ \log p_J[s=\mathbf 1] \big] \Big),
\end{equation}
and the two coincide to leading order in the disorder and whenever $\mathrm{SAR}$ is close to one (\cref{app:relation}); both are evaluated in the numerics of \cref{app:numerics}.
To leading order in the coupling strength (\cref{app:smallJ}),
\begin{equation}\label{eq:leading}
    \log \mathrm{SAR}_{\rm geo} = -N\log\!\big(1+e^{-2h}\big) - \frac{J_0^2}{4}\,N(N-1)\big(1-\tanh^4 h\big) + \mathcal{O}(J_0^4)
\end{equation}
(we have verified this coefficient against exact enumeration, \cref{app:smallJ}). The disorder thus contributes a term quadratic in $N$ with a fixed negative sign, which is precisely the quadratic law $\exp(-aN-bN^{2})$ compared in \cref{sec:comparison}, with $a=\log(1+e^{-2h})$ and $b\simeq 2J_0^2e^{-2h}$ for $h\gg1$. The empirical form \cref{emp_sar} corresponds to exponentiating the relative correction, $1+2J_0^2(N-1)\to e^{2J_0^2(N-1)}$, which suggests the identification $\alpha\simeq e^{2J_0^2}$, $\beta_0\simeq e^{-2h}$. This exponentiation is a phenomenological closed form rather than a controlled resummation, and we do not use it for quantitative statements. Instead, we evaluate the model directly by exhaustive enumeration of the $2^N$ configurations for $N\le19$ and $200$ disorder realizations (\cref{app:numerics}, \cref{fig:sk-numerics}). The exact curves display the qualitative phenomenology of the data. A plateau near $\mathrm{SAR}=1$ is followed by a crossover whose location is controlled by $J_0$ and $h$ and which is much sharper than the exponential law with the same $h$. The crossover of the exact curves is, however, broader than the exponentiated form predicts, and the leading-order quadratic law is accurate only well before the crossover. We therefore regard the model as a \emph{minimal effective model that qualitatively produces super-linear error accumulation and an accuracy cliff}.

From the statistical-physics perspective, the accuracy cliff corresponds to a finite-size crossover in the underlying spin system. For small system size $N$, the external field $h$ dominates, and the system behaves as a paramagnet, aligning most spins in the correct direction ($s_i = +1$). As $N$ increases, the disorder energy of the random couplings, which by construction scales as $N^2$, eventually exceeds the field and entropic contributions that grow only linearly with $N$. Beyond a characteristic size $N_*$ the fully aligned configuration is no longer favored, leading to the accuracy degradation observed as the accuracy cliff.

\begin{figure}[t]
    \centering
    \includegraphics[width=0.85\linewidth]{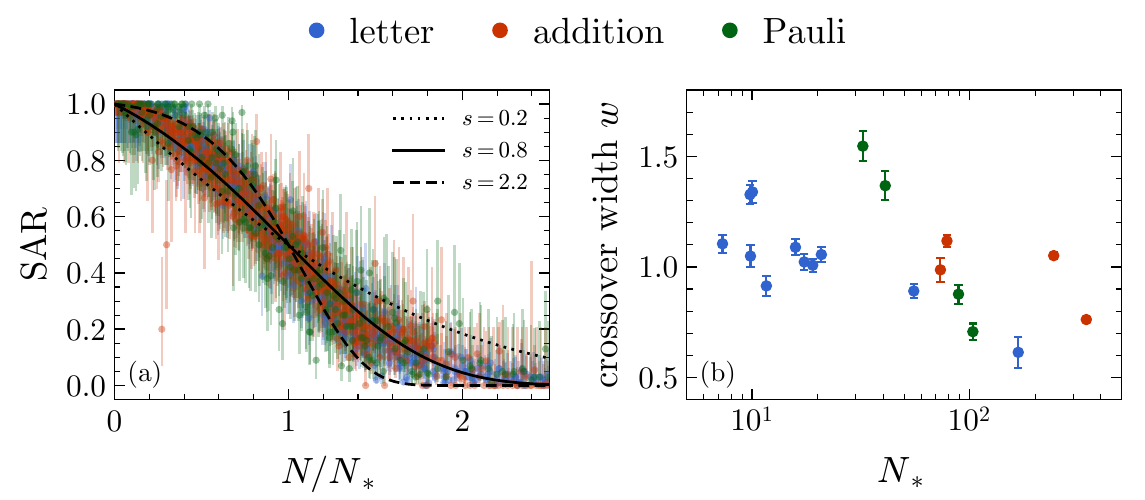}
    \caption{Scaling collapse of the Sequence Accuracy Rate. (a) SAR against the rescaled length $N/N_*$ (with $N_*$ the fitted $\mathrm{SAR}=0.5$ crossover scale of \cref{emp_sar}) for all 19 model--task curves; colors denote task families, error bars are $95\%$ Wilson intervals with the batch design effect. The black curves are the rescaled form of \cref{emp_sar}, $\exp[-(\ln2)x\,e^{s(x-1)}]$, for shape parameters $s=N_*\log\alpha=0.2$, $0.8$ and $2.2$ spanning the fitted range. (b) The law-independent crossover width $w=[N(\mathrm{SAR}{=}0.2)-N(\mathrm{SAR}{=}0.8)]/N_*$ of every curve against its $N_*$, with standard errors; the widths differ significantly between model--task pairs.}
    \label{fig:data-collapse}
\end{figure}

\paragraph{Scaling collapse.} A crossover governed by a single scale suggests plotting SAR against $N/N_*$. \cref{fig:data-collapse}(a) shows that the data of all model--task pairs then fall within a common band. Its scatter about a non-parametric master curve (RMS $0.068$) equals the scatter about each curve's own fit ($0.072$) and the typical statistical error ($0.071$). This collapse should not, however, be read as evidence of a universal scaling function. Because $N_*$ is fitted for each curve, every curve passes through $(1,\tfrac12)$ by construction, and the rescaled form of \cref{emp_sar}, $\mathrm{SAR}(xN_*)=\exp[-(\ln2)\,x\,\alpha^{N_*(x-1)}]$, retains one shape parameter $s=N_*\log\alpha$. Across our 19 curves $s$ ranges from $0.21$ to $2.15$, and the law-independent crossover width $w=[N(\mathrm{SAR}{=}0.2)-N(\mathrm{SAR}{=}0.8)]/N_*$ from $0.61$ to $1.55$ with standard errors of about $0.04$ (\cref{fig:data-collapse}(b)); the variation of $w$ is highly significant ($\chi^2=9\times10^{2}$ for 18 degrees of freedom). Rescaling instead with the $N_*$ of the quadratic or of the stretched-exponential law gives the same scatter about the corresponding master curve ($0.068$), so the collapse does not discriminate between crossover laws either. The collapse therefore holds at the level of the statistical noise, but the fitted widths differ significantly between model--task pairs, so the master curve is common only approximately.

\begin{figure}[t]
    \centering
    \includegraphics[width=0.95\linewidth]{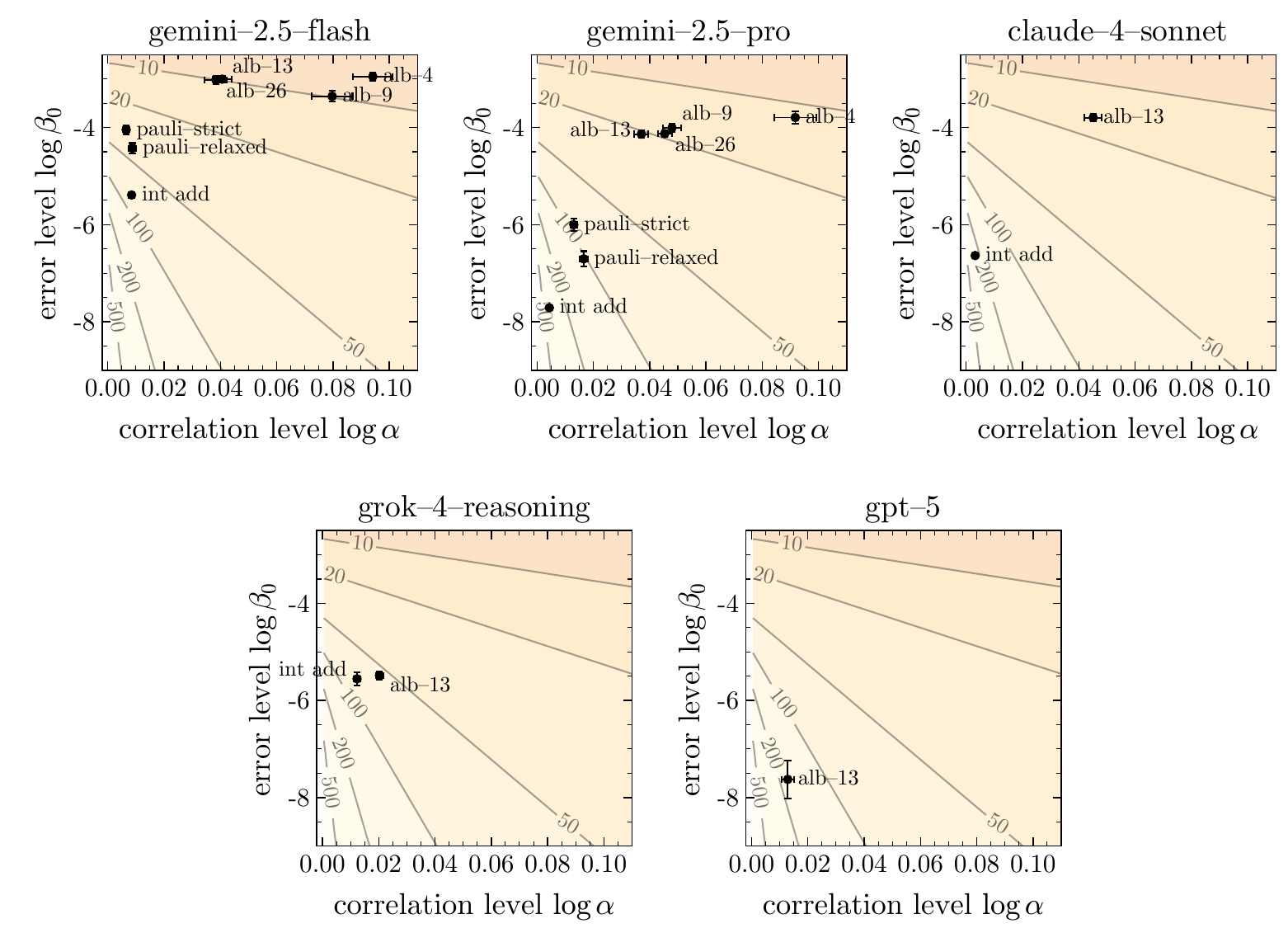}
    \caption{Correlation–error map grouped by model. Each point is the maximum-likelihood $(\log \alpha, \log \beta_0)$ of \cref{emp_sar} for one task, with $1\sigma$ error bars (quasi-likelihood); the horizontal and vertical axes represent the correlation level and error level, respectively. The background color and the gray contours give the crossover scale $N_*$ implied by \cref{emp_sar} at each point (numbers along the contours).}
    \label{fig:phase-model}
\end{figure}

\begin{figure}[t]
    \centering
    \includegraphics[width=0.65\linewidth]{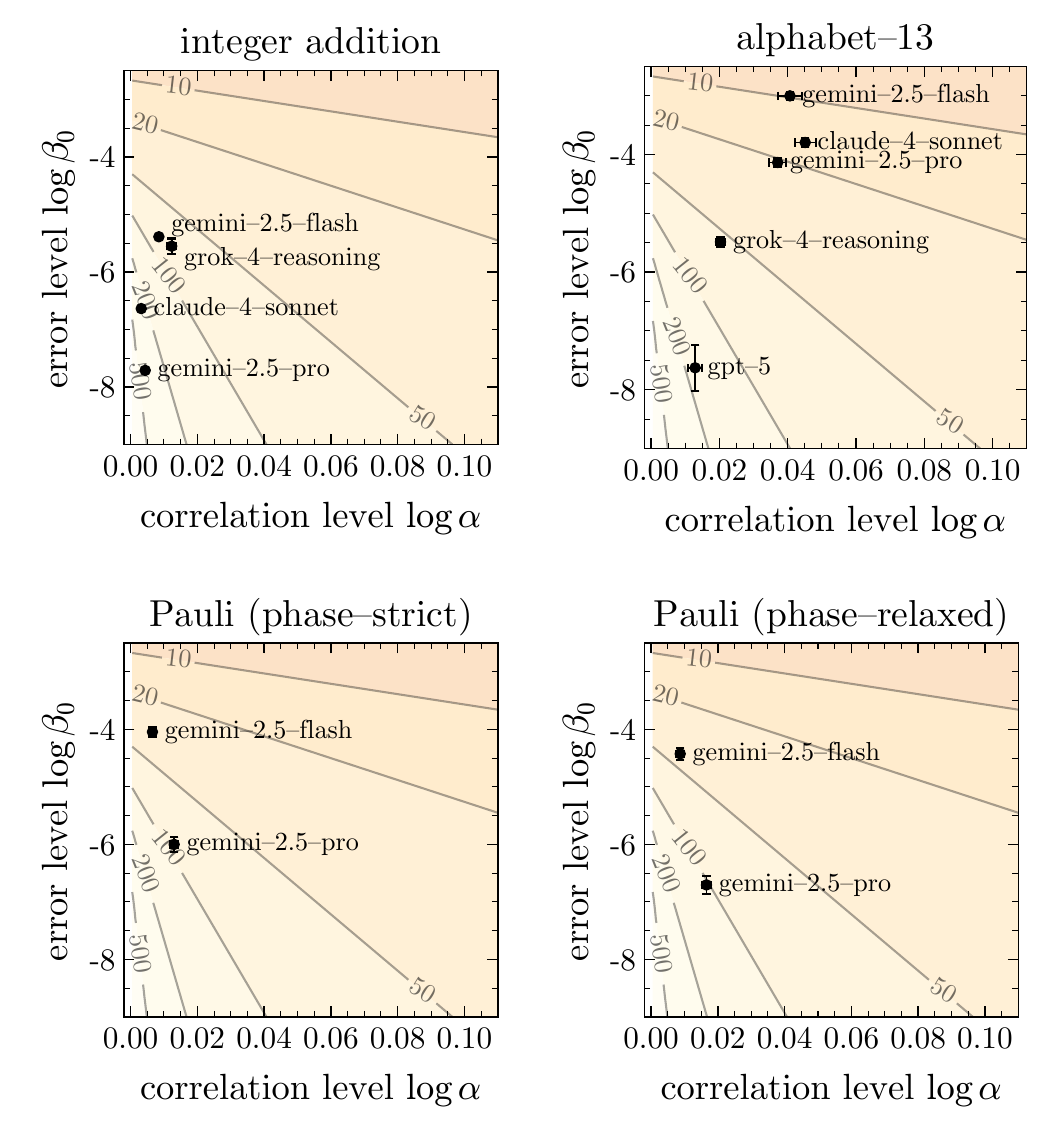}
    \caption{Correlation–error map grouped by task; conventions as in \cref{fig:phase-model}.}
    \label{fig:phase-task}
\end{figure}

\paragraph{Correlation--error map.} Using the fitted parameters of \cref{emp_sar}, we pin each model--task pair onto a correlation--error map. Each point corresponds to a pair $(\log \alpha, \log \beta_0)$, with the crossover scale $N_*$ (the solution of $\mathrm{SAR}(N_*) = 0.5$) shown as background contours. Within this parametrisation, a smaller $\log\alpha$ means that errors accumulate less strongly along the output, and a smaller $\log\beta_0$ means a lower error rate for executing a single unit task; since the two parameters are correlated in the fit and $\alpha$ is form-dependent (\cref{sec:comparison}), the robust content of the map is the position of each pair relative to the $N_*$ contours.
The mapping results grouped by model are shown in \cref{fig:phase-model},
and those grouped by task are shown in \cref{fig:phase-task}.

\section{Implication: Divide-and-Conquer Strategy}

The empirical scaling law of Eq.~(\ref{emp_sar}) not only provides a qualitative description of how the accuracy inevitably degrades with the output sequence length $N$ for LLMs, but also offers a \emph{constructive implication}: when $\alpha > 1$, the reliability of long-sequence generation can be improved by adopting a \emph{divide-and-conquer} strategy. We emphasize that the prediction below follows directly from the empirical accumulation law \cref{emp_sar}; it does not depend on the effective model of \cref{sec:model}, and the experiment in this section is presented as a practical implication of the accumulation law, not as a test of the statistical-mechanical picture.

In this context, divide-and-conquer refers to partitioning a sequence-generation task of total length $N$ into $k$ sub-tasks of length $N/k$, prompting the model separately for each sub-task, and then merging the partial outputs to form the complete sequence. The intuition is that the accuracy cliff originates from correlated error propagation along the output: when $\alpha>1$, the effective per-position error rate grows exponentially with $N$. Dividing the sequence into shorter fragments resets this accumulation.

We introduce a multiplicative overhead factor $\theta_{N,k}\in(0,1]$ to account for the reduction of SAR due to the additional operations involved in applying the divide-and-conquer strategy, such as segmentation, formatting, boundary handling, and content copying. The overall success probability under divide-and-conquer is then
\begin{equation}\label{eq:SAR_DC}
\mathrm{SAR}_{\mathrm{DC}}(N,k)
= \theta_{N,k}\,\big[\mathrm{SAR}(N/k)\big]^k
= \theta_{N,k}\,\exp\!\Big[-\beta_0\,N\,\alpha^{\,\frac{N}{k}-1}\Big].
\end{equation}
Define the logarithmic gain of divide-and-conquer relative to the single-shot strategy as
\begin{equation}
\Delta(N,k)
:= \log \frac{\mathrm{SAR}_{\mathrm{DC}}(N,k)}{\mathrm{SAR}(N)}
= \log \theta_{N,k} + \beta_0 N \Big( \alpha^{\,N-1} - \alpha^{\,\frac{N}{k}-1} \Big).
\end{equation}
This formulation shows that for $\alpha>1$, the gain term increases rapidly with $N$, indicating that the larger $\alpha$ is, the more advantageous segmentation becomes.

\begin{theorem}[Advantage of Divide-and-Conquer]\label{theorem1}
Let the intrinsic error rate $\beta_0>0$ and the correlation amplification factor $\alpha>1$, and fix the integer $k\ge2$.
Under the empirical scaling law $\mathrm{SAR}(N)=\exp[-\beta_0 N \alpha^{\,N-1}]$, for the segmentation into $k$ equal sub-tasks to yield a positive gain $\Delta(N,k)>0$, it would be sufficient if the sequence length exceeds the following bound
\begin{equation}
N\geq N_{\mathrm{DC}} \;=\; 1 + \frac{1}{\log\alpha}\left(\log\left(1-\frac{2\log\theta_{N,k}}{\beta_0}\right)+\frac{\log 2}{1-1/k}\right),
\end{equation}
meaning that for all $N \ge N_{\mathrm{DC}}$, one has $\mathrm{SAR}_{\mathrm{DC}}(N,k) > \mathrm{SAR}(N)$.
\end{theorem} (See App.~\ref{app:proof} for a proof.)

This result provides an explicit criterion for when the divide-and-conquer approach is beneficial.
Under this strategy, as evident from Eq.~\eqref{eq:SAR_DC}, the intrinsic error rate $\beta_0$ remains unchanged, while the correlation amplification factor $\alpha$ is effectively rescaled to $\alpha_\text{eff}\simeq \alpha^{1/k}$ in the large-$N$ regime.
As a consequence, the characteristic accuracy-cliff scale is extended to
\begin{equation}
    N_* \simeq 1+\frac{\log(1/\beta_0)}{\log \alpha_\text{eff}}=1+k\,\frac{\log(1/\beta_0)}{\log \alpha},
\end{equation}
growing approximately linearly with $k$, according to Eq.~\eqref{eq:Nc}.

\begin{figure}[t]
    \centering
    \includegraphics[width=0.95\linewidth]{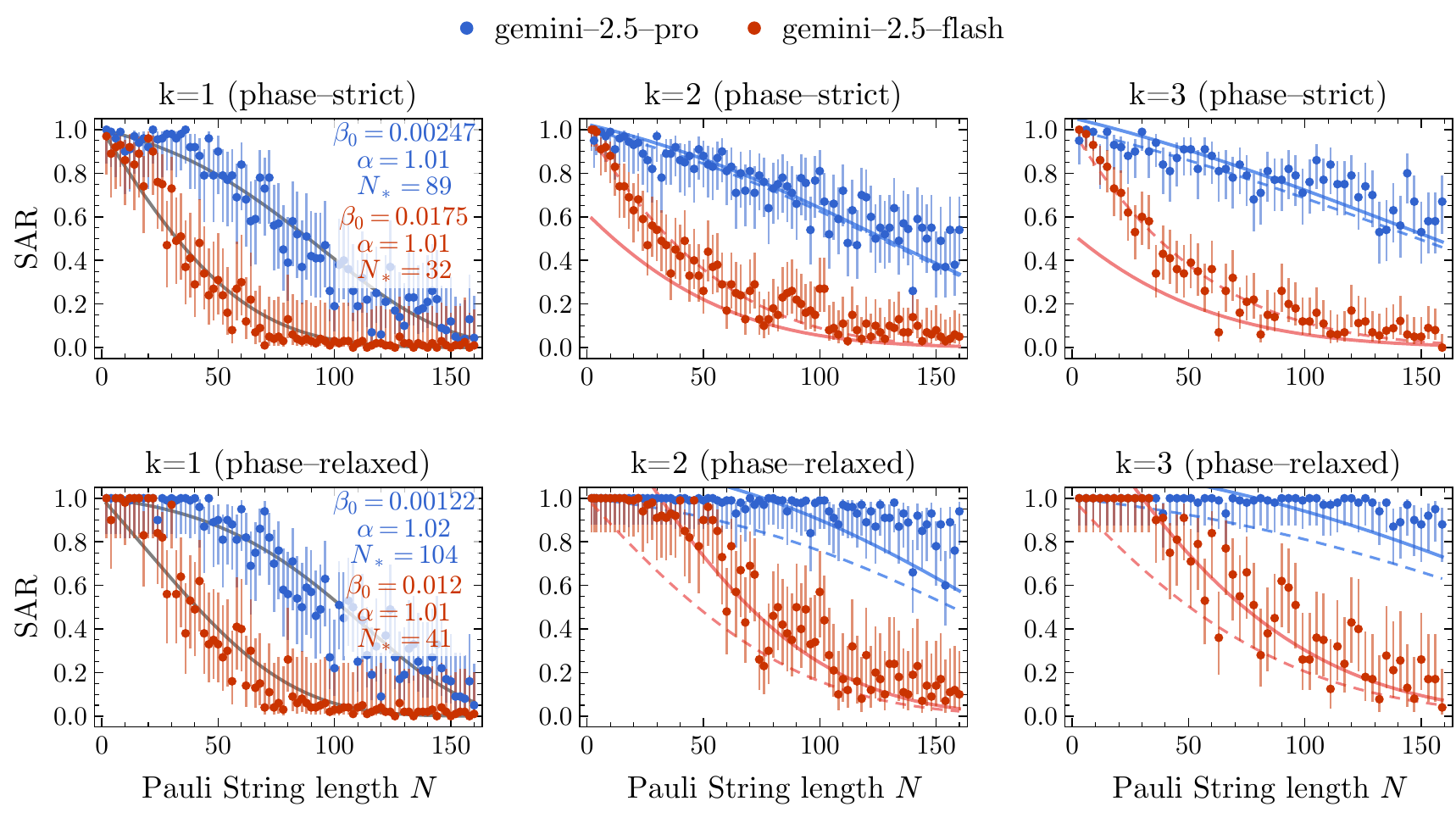}
    \caption{
Divide-and-conquer evaluation on the Pauli string multiplication task under the phase-strict (top) and phase-relaxed (bottom) criteria, using \texttt{gemini-2.5-pro} (blue) and \texttt{gemini-2.5-flash} (red). Points: fraction of correct instances with $95\%$ intervals. The gray $k{=}1$ curves are the maximum-likelihood fits of \cref{emp_sar}; the extracted $(\alpha,\beta_{0})$ are then used to draw $\mathrm{SAR}(N/k)^{k}$ for $k{=}2$ and $k{=}3$ without overhead (dashed) and multiplied by the overhead measured from the same records (solid; the product of the chunk-stage and recombination factors quoted in the text). For \texttt{pro} the measured overhead is negligible and the data follow the prediction semi-quantitatively; for \texttt{flash} the recombination overhead ($\theta=0.82$, $0.76$) and a chunk-stage shortfall suppress the plateau under the strict criterion.}\label{fig:d&c}
\end{figure}

\paragraph{Measurement.} We test this implication on the Pauli string multiplication task, whose sub-problems are exactly site-local and whose recombination is unambiguous (\cref{fig:d&c}); extending the test to the other task families is left to future work. Because every stage of the pipeline is recorded, the overhead is not a free parameter: from the same records we measure the fraction of problems that were decomposed faithfully, the fraction whose $k$ chunk products were all correct, and the probability that the final recombination is correct given correct chunk products, which is the overhead factor $\theta$ of \cref{eq:SAR_DC}. For \texttt{gemini-2.5-pro}, decomposition is faithful in $99\%$ of the problems and $\theta=0.998$ ($k=2$) and $0.995$ ($k=3$) under the strict criterion, i.e.\ the overhead is negligible; for \texttt{gemini-2.5-flash}, decomposition is faithful in $98\%$ ($k=2$) but only $76\%$ ($k=3$) of the problems and $\theta=0.82$ ($k=2$) and $0.76$ ($k=3$), so the weaker model loses a substantial fraction of its correct chunk products in the recombination step. Under the phase-relaxed criterion $\theta=1.00$ for both models. The chunk stage itself is only semi-quantitatively described by $[\mathrm{SAR}(N/k)]^k$ with the single-shot fit. Under the strict criterion the measured probability that all chunks are correct exceeds this prediction by $3$--$6\%$ for \texttt{pro} and falls short of it by $25$--$31\%$ for \texttt{flash}, so \cref{eq:SAR_DC} with a constant $\theta$ is a guide rather than a precise prediction. The observation that survives these caveats is the extension of the crossover scale. Reading $N_*$ directly from the data (\cref{fig:d&c}), \texttt{pro} moves from $N_*\approx80$ ($k=1$) to $136$ ($k=2$) and $155$ ($k=3$) under the strict criterion, and from $94$ to beyond the measured range ($N>160$) under the relaxed criterion; \texttt{flash} moves from $39$ to $71$ and $77$ under the relaxed criterion, whereas under the strict criterion it shows no plateau for $k\ge2$, consistent with its recombination overhead.

In summary, the empirical scaling law does more than describe the accuracy cliff; it also indicates how to mitigate it. Segmentation reduces the compounding of correlated errors and extends the \emph{reliable length scale} of a sufficiently capable model by a factor comparable to $k$, at the price of an overhead that is model-dependent and can be measured.



\section{Related Works}

\textbf{Single-pass reasoning and capacity limits.}
Recent research has shown that even when models can locally apply correct reasoning steps, global accuracy collapses as reasoning chains lengthen. This phenomenon parallels information-theoretic capacity limits of single forward passes. The \emph{Fano-style upper bound} framework ~\citep{wan2025fano} formalizes this collapse for multi-hop question answering (MHQA), proving that once task complexity exceeds the model’s per-pass information capacity, accuracy must drop sharply—mirroring our empirical SAR crossover. Their work interprets the reasoning chain as a noisy communication channel, bounded by mutual information, while our SAR analysis characterises the same breakdown empirically and offers an effective statistical-physics description in which correlated error propagation along the output plays the central role. Together they highlight the inevitability of reliability loss in single-pass reasoning under finite model capacity.

\textbf{Prompting paradigms and decomposition.}
A wide range of prompting methods—Direct prompting, Chain-of-Thought (CoT)~\citep{wei2022chain,kojima2022large}, ReAct~\citep{yao2023react}, Plan-and-Solve~\citep{wang2023plan}, and Self-Ask~\citep{press2023measuring}—operate in the \emph{single-pass} regime, assuming the full reasoning chain fits within one generation. By contrast, \emph{multi-call} frameworks such as Self-Refine~\citep{madaan2023self}, InfoQA~\citep{wan2025fano}, and agentic systems for programming or reasoning~\citep{li2024survey,qian2024chatdev,kim2024llm} mitigate this bottleneck by decomposing reasoning into smaller, verifiable subcalls. Our SAR-based analysis provides a quantitative explanation of why such multi-call methods succeed: by resetting the error accumulation process, they effectively bound per-step complexity below the model’s capacity threshold, preventing the exponential or super-exponential degradation seen in single-pass reasoning.

\textbf{Self-checking, verification, and robust reasoning.}
Other orthogonal strategies—such as self-consistency sampling, debate-style verification, and tool-augmented reasoning—attempt to mitigate compounding errors by introducing redundancy or external validation~\citep{yao2023react,xie2025fire,wan2025cognitive,mccoy2023embers}. SAR offers a unified diagnostic lens for such methods: improved reliability corresponds to reduced intrinsic error rate $\beta_0$, while better verification or decomposition corresponds to smaller accumulation factor $\alpha$. Thus, SAR quantitatively links algorithmic interventions to empirical stability regimes.

\section{Conclusion and Future Work}

We introduced the \emph{Sequence Accuracy Rate (SAR)} as a principled metric for quantifying reliability in deterministic reasoning tasks performed by large language models. Through three controlled benchmarks—cyclic letter replacement, integer addition, and Pauli string multiplication—we demonstrated that SAR provides a quantitative view of how local inaccuracies compound with sequence length. Empirical results reveal a reliability crossover that is sharper than a simple exponential decay; the independent-error law is rejected on every model--task pair, which indicates correlated rather than independent error accumulation. The scale $N_*$ of the crossover is robustly determined, independently of the functional form used to fit it, while its shape varies moderately between model--task pairs.

To interpret this behavior, we introduced a minimal effective model in which task-position correctness variables interact through dense random couplings of Sherrington--Kirkpatrick type and compete with an external field set by the prompt. Evaluated by direct enumeration, the model reproduces super-linear error accumulation and an accuracy cliff qualitatively, and it provides a two-parameter description of each model--task pair in terms of an intrinsic error rate $\beta_0$ and an error accumulation factor $\alpha$. We regard it as a phenomenological picture of correlated error accumulation.

Future directions include extending SAR to stochastic reasoning pipelines, integrating it with self-verification or multi-call decomposition schemes, and applying the statistical-physics approach to study model scaling laws and architectural modifications. Two concrete next steps are (i) extending the divide-and-conquer test beyond Pauli string multiplication to the cyclic-letter-replacement and integer-addition tasks, (ii) making the connection to network internals microscopic---for instance, by attempting to extract the effective couplings $J_{ij}$ from attention patterns, or testing whether controlled perturbations of attention sparsity shift the accumulation factor $\alpha$ in the predicted direction; and (iii) studying a causal variant of the effective model with couplings only from earlier to later positions. More broadly, SAR provides a foundation for treating reasoning reliability as a measurable and scalable property, enabling systematic analysis and engineering of models that sustain compositional accuracy over increasing complexity.

\section{Acknowledgment}
We would like to thank Chen Nie for the engineering support. We would like to acknowledge helpful discussion with Victor Albert and Hsin-Yuan Huang on \href{https://bench.science}{bench.science}. W.H. additionally thanks Tianxiao Hu for helpful discussions.

\bibliography{ref}
\bibliographystyle{plainnat}

\newpage
\appendix
\crefalias{section}{appendix}
\crefalias{subsection}{appendix}
\crefalias{subsubsection}{appendix}

\section{Success Rate Analysis of Pauli String test}\label{app:pauli_sar}

Suppose an intellectual agent attempts to compute the product of two Pauli strings of length $N$.
At each site, two types of operations are required:

\begin{enumerate}
    \item A \emph{Pauli multiplication step}, where $P_i \times Q_i$ is reduced to a Pauli operator $R_i$ and a local phase factor $\phi_i$. Let the probability of correctly performing this step be $p_{\sigma}$.
    \item A \emph{phase multiplication step}, where the accumulated phase is updated as $\Phi \mapsto \Phi \times \phi_i$. Let the probability of correctly performing a single phase update be $p_{\phi}$.
\end{enumerate}

We first model the accumulated phase as a Markov chain on the cyclic group
\(G = \{1, i, -1, -i\}\) with the state order \((1, i, -1, -i)\).
At each step, the algorithm multiplies the current phase by one of the group elements,
but due to noise it returns an incorrect element with probability $p_{\phi}$,
and the correct (true) element with probability \(1 - p_{\phi}\).
Hence, each of the three incorrect elements occurs with probability \(\tfrac{p_{\phi}}{3}\).
Let \(\mathds{1}\) denote the \(4\times4\) identity matrix and \(J\) the all-ones matrix.
When the intended multiplier is \(1\), the one-step transition kernel is the circulant matrix
\begin{equation}
    K = (1 - p_{\phi})\,\mathds{1} + \frac{p_{\phi}}{3}\,(J - \mathds{1}).
\end{equation}

Define the four permutation (left-shift) matrices \(\{S_\phi:\phi\in G\}\) that implement left multiplication by \(\phi\) on the state order; explicitly,
\[
S_1=\mathds{1},\qquad
S_i=\begin{bmatrix}
0&0&0&1\\
1&0&0&0\\
0&1&0&0\\
0&0&1&0
\end{bmatrix},\qquad
S_{-1}=\begin{bmatrix}
0&0&1&0\\
0&0&0&1\\
1&0&0&0\\
0&1&0&0
\end{bmatrix},\qquad
S_{-i}=\begin{bmatrix}
0&1&0&0\\
0&0&0&1\\
0&0&1&0\\
1&0&0&0
\end{bmatrix}.
\]

Thus, if the multiplier at a step is \(\phi\in G\), the corresponding one-step transition matrix is
\begin{equation}
T(\phi)\;=\; S_\phi\, K.
\end{equation}

Because every permutation matrix commutes with $K$, i.e.\ $S_\phi K = K S_\phi, \forall \phi \in G$,
the product of step-wise transitions reduces to a similarity transform. For any sequence of true multipliers $\phi_1,\ldots,\phi_N$, we have
\begin{equation}
T(\phi_N)\cdots T(\phi_1)
= \big(S_{\phi_N}\cdots S_{\phi_1}\big)\, K^{N}
= S_{\Phi}\, K^{N},
\quad \Phi:=\phi_N\cdots\phi_1.
\end{equation}
Therefore, the success rate of phase computing task is given by:

\begin{equation}
    P_{\phi}(N)=e_{\Phi}^{\top}S_{\Phi}\, K^{N}e_{1},
\end{equation}

where $e_{\Phi}$ is the basis vector of $\Phi$ in the state order $(1,i,-1,-i)$ and $K^{N}=\tfrac{1}{4}J+\Big(\tfrac{3-4p_{\phi}}{3}\Big)^{N} (\mathds{1}-\tfrac{1}{4}J)$. Explicitly, this gives:

\begin{equation}
P_{\phi}(N)
= \tfrac14+\tfrac34\Big(\tfrac{3-4p_{\phi}}{3}\Big)^{N}.
\end{equation}

Combining this with the individual Pauli multiplication success rate, the overall sequence accuracy rate is
\begin{equation}
\text{SAR}(N) = (1-p_{\sigma})^{N}\left[\tfrac14+\tfrac34\left(\tfrac{3-4p_{\phi}}{3}\right)^{N}\right].
\end{equation}

The requirement to accumulate local phase factors across all sites is especially instructive, as it directly reflects the need for \emph{memory handling} in agent design: just as an LLM must correctly maintain and update a state variable across multiple reasoning steps, the phase register here must be updated consistently at every site.

\section{An effective random-coupling model for the SAR}\label{app:sk-sar}

In this appendix, we define and analyse the minimal energy-based model of \cref{sec:model}: its perturbative expansion to leading order in the coupling (\cref{app:smallJ}), its direct evaluation by exhaustive enumeration (\cref{app:numerics}), and the relation between the empirical estimator of SAR and the disorder averages used in the model (\cref{app:relation}).

Let \(\mathcal{D} = \{(x_{(k)}, y_{(k)})\}_{k=1}^{K}\) be a dataset consisting of input–output sequence pairs,
where each input sequence \(x_{(k)}\) is deterministically mapped to a unique correct output sequence \(y_{(k)}\).
An LLM trained on this dataset defines a conditional probability distribution
\(p_{\theta}(y \mid x)\) over output sequences.

\subsection{Definitions and independent-token baseline}\label{app:def}
The \emph{Sequence Accuracy Rate (SAR)} is defined as the geometric mean of the model's probability on correct sequences:
\begin{equation}
\mathrm{SAR}(p_\theta;\mathcal{D})
=\exp\!\Big(\frac{1}{K}\sum_{k=1}^{K}\log p_\theta(y_{(k)}\mid x_{(k)})\Big)\in(0,1].
\end{equation}

Introduce Ising variables for correctness:
\[
s_i=\begin{cases}
+1,& y_i\ \text{is correct},\\
-1,& \text{otherwise},
\end{cases}
\qquad s=(s_1,\dots,s_N)\in\{\pm1\}^N,
\]
so all-correct output sequence corresponds to \(s_i = +1\) for all \(i = 1, \ldots, N\),
which we denote jointly as \(s = \mathbf{1}\).
If each token were generated independently with a token-level accuracy rate \(p_{\mathrm{tok}}\),
the expected sequence accuracy would scale as
\begin{equation}
\mathrm{SAR} = p_{\mathrm{tok}}^{N},
\end{equation}
which decays exponentially with sequence length \(N\).
The corresponding energy model for the Ising variables is simply
\begin{equation}
E[s] = -h \sum_{i=1}^{N} s_i,
\end{equation}
such that
\begin{equation}
\mathrm{SAR}
= \frac{e^{-E[s=\mathbf{1}]}}{\sum_{s} e^{-E[s]}}
= \left(\frac{1}{1 + e^{-2h}}\right)^{N},
\end{equation}
implying that the external field \(h\) parameterizes the token-level accuracy as
\begin{equation}
p_{\mathrm{tok}} = \frac{1}{1 + e^{-2h}}.
\end{equation}
However, this independent-position prediction fails to capture
the crossover behavior observed in experiments.

\paragraph{Empirical estimator versus model averages.}\label{app:relation}
The quantity measured in \cref{sec:experiment} is the fraction of problem instances whose complete answer is correct. Its expectation is the \emph{arithmetic} instance average of the per-instance success probability, $\mathbb{E}_x\,p_\theta(y_x\mid x)$, where $y_x$ is the unique correct answer to instance $x$. In the effective model, problem instances are represented by disorder realisations of the couplings, so the model counterpart of the empirical SAR is the arithmetic disorder average $\mathrm{SAR}_{\rm ari}=\mathbb{E}_J\,p_J[s=\mathbf 1]$. The replica calculation below is most naturally carried out for the geometric (``quenched'') average $\mathrm{SAR}_{\rm geo}=\exp(\mathbb{E}_J\log p_J[s=\mathbf 1])$ of \cref{ising_sar}. The two averages coincide to leading order in $J_0$ (the disorder variance of $\log p_J$ enters only at $\mathcal O(J_0^2 e^{-4h})$) and whenever $\mathrm{SAR}$ is close to one; through the crossover the arithmetic average is larger, because it is dominated by favourable realisations, and the two are compared numerically in \cref{app:numerics}. Neither average is the quantity defined at the beginning of this subsection (a geometric mean over instances of the model's probability of the correct answer), which is a property of the language model's output distribution rather than of the binary correctness outcomes; we retain that definition only as motivation.

\subsection{Energy model (dense random couplings of Sherrington--Kirkpatrick type)}\label{app:sk}

In our experiments, the Sequence Accuracy Rate (SAR) shows a pronounced dependence on output length \(N\).
While one might expect a simple exponential decay, \(\mathrm{SAR} \sim e^{-\alpha N}\), our results reveal a clear \emph{accuracy cliff}: SAR remains high for short sequences, then sharply declines around a characteristic length.
This behavior indicates that sequence-level reliability in LLMs arises not from independent errors but from correlated effects governing the shift from reliable to unreliable generation in deterministic tasks.
We hypothesize that the deviation arises from dense correlations among the output positions during sequence generation.
Because attention couples tokens across the sequence, dependencies beyond the independent-position picture can arise; we capture these in an effective way, without claiming a specific microscopic origin.
We represent them by effective random coupling strengths \(J_{ij}\) between the Ising variables \(s_i\) (the position-correctness labels), whose random signs encode constructive and destructive interference.
Such couplings modify the collective statistics of sequence correctness, producing a non-exponential decay and the observed crossover in SAR.

We therefore propose the following energy model:
\begin{equation}
E_J[s] = -\sum_{i<j} J_{ij}\, s_i s_j - h \sum_{i=1}^{N} s_i,
\end{equation}
where \(s_i = \pm 1\) for \(i = 1, 2, \ldots, N\) are Ising variables representing the correctness of the \(i\)-th position \(y_i\) of the output generated by the LLM.

The coupling strengths \(J_{ij}\) are drawn independently from a Gaussian distribution with zero mean and finite variance:
\begin{equation}
\mathbb{E}_J[J_{ij}] = 0,
\qquad
\mathbb{E}_J[J_{ij}^2] = J_0^2.
\end{equation}
They model the noisy all-to-all correlations between output positions, and \(J_0\) characterizes the noise strength. The variance \(J_0^2\) is taken independent of \(N\) (see the discussion in \cref{sec:model}); with the conventional SK normalisation \(J_0^2/N\) the disorder energy would scale like the field energy and, as shown in \cref{app:numerics}, no accuracy cliff would arise.
The external field \(h\) parameterizes the position-level accuracy rate via \(p_{\mathrm{tok}}=1/(1+e^{-2h})\): a strong field \(h\) corresponds to a high position-level accuracy, which is typically assumed. The energy function is that of the \emph{Sherrington--Kirkpatrick (SK) model} of spin-glass physics, up to the normalisation of the couplings.

For a particular random realization of \(J_{ij}\), the probability of observing the Ising sequence \(s\) is
\begin{equation}
p_J[s] = \frac{1}{Z_J} \, e^{-E_J[s]},
\end{equation}
where \(Z_J\) is the partition function,
\begin{equation}
Z_J = \sum_{s} e^{-E_J[s]}.
\end{equation}

In the perturbative analysis we use the geometric disorder average of the probability of the all-correct configuration (\cref{app:relation}):
\begin{equation}\label{eq:logSAR-master}
\mathrm{SAR}_{\rm geo} := \exp\!\big(\mathbb{E}_J \log p_J[s=\mathbf{1}]\big).
\end{equation}

\subsection{Replica trick and disorder average}\label{app:replica}

From \cref{eq:logSAR-master}, it is straightforward to show that
\begin{align}
\log \mathrm{SAR}
&= \mathbb{E}_J \log p_J[s=\mathbf{1}]
= \mathbb{E}_J \log\!\left(\frac{e^{-E_J[s=\mathbf{1}]}}{Z_J}\right) \nonumber\\
&= \mathbb{E}_J \big(-E_J[s=\mathbf{1}] - \log Z_J\big)
= \mathbb{E}_J\!\Big(\sum_{i<j} J_{ij} + hN - \log Z_J\Big).
\end{align}
Using the disorder mean \(\mathbb{E}_J[J_{ij}] = 0\), we have
\begin{equation}
\label{eq:log-sar-basic}
\log \mathrm{SAR}
= hN - \mathbb{E}_J \log Z_J .
\end{equation}

\paragraph{Replica trick.}
The expectation \(\mathbb{E}_J \log Z_J\) in \eqref{eq:log-sar-basic} is intractable directly.
Physicists therefore use the \emph{replica trick}:
\begin{equation}
\label{eq:replica-identity}
\mathbb{E}_J \log Z_J
= \lim_{n\to 0} \partial_n \, \mathbb{E}_J \big[ Z_J^{\,n} \big].
\end{equation}
By definition,
\begin{align}
\mathbb{E}_J \big[ Z_J^{\,n} \big]
&= \mathbb{E}_J \left[
\Big(\sum_{[s]} \exp\big(\sum_{i<j} J_{ij} s_i s_j + h \sum_i s_i\big)\Big)^{\!n}
\right] \nonumber\\
&= \sum_{\{s^a\}} \exp\!\Bigg(
\sum_{a=1}^n \sum_{i<j} J_{ij}\, s_i^a s_j^a
+ h \sum_{a=1}^{n} \sum_{i=1}^{N} s_i^a
\Bigg),
\end{align}
where the spin variables have been replicated to \(s_i^a\) with \(i=1,\ldots,N\) and \(a=1,\ldots,n\).
Here \(N\) is the number of sites (sequence length) and \(n\) the number of replicas.

\paragraph{Disorder average.}
Averaging \(Z_J^{\,n}\) over i.i.d.\ Gaussian couplings \(J_{ij}\) with
\(\mathbb{E}_J[J_{ij}]=0\), \(\mathbb{E}_J[J_{ij}^2]=J_0^2\), we obtain
\begin{align}
\mathbb{E}_J \big[ Z_J^{\,n} \big]
&= \sum_{\{s^a\}}
\prod_{i<j}
\mathbb{E}_{J_{ij}}
\exp\!\Big(\sum_{a=1}^{n} J_{ij}\, s_i^a s_j^a\Big)\,
\exp\!\Big(h \sum_{a=1}^{n} \sum_{i=1}^{N} s_i^a\Big) \nonumber\\
&= \sum_{\{s^a\}}
\prod_{i<j}
\exp\!\Big(\frac{J_0^2}{2}\, C_{ij}^2\Big)\,
\exp\!\Big(h \sum_{a=1}^{n} \sum_{i=1}^{N} s_i^a\Big),
\end{align}
where \(C_{ij} \equiv \sum_{a=1}^{n} s_i^a s_j^a\) and we used the Gaussian moment-generating function
\(
\mathbb{E}_{J_{ij}} e^{J_{ij} C_{ij}}
= \exp\!\big(\tfrac{J_0^2}{2} C_{ij}^2\big).
\)
Thus,
\begin{align}
\mathbb{E}_J \big[ Z_J^{\,n} \big]
&= \sum_{\{s^a\}}
\exp\!\Big(
\frac{J_0^2}{2} \sum_{i<j} C_{ij}^2
+ h \sum_{a=1}^{n} \sum_{i=1}^{N} s_i^a
\Big) \nonumber\\
&= \sum_{\{s^a\}}
\exp\!\Big(
\frac{J_0^2}{2} \sum_{i<j} \big(\sum_{a=1}^{n} s_i^a s_j^a\big)^2
+ h \sum_{a=1}^{n} \sum_{i=1}^{N} s_i^a
\Big).
\end{align}

Expanding the quadratic replica term gives
\begin{align}
\sum_{i<j} \Big(\sum_{a=1}^{n} s_i^a s_j^a\Big)^2
&= \sum_{i<j} \left(\sum_{a=1}^{n} (s_i^a s_j^a)^2
+ 2 \sum_{a<b} s_i^a s_j^a s_i^b s_j^b \right) \nonumber\\
&= \sum_{i<j} \left(n + 2 \sum_{a<b} s_i^a s_j^a s_i^b s_j^b \right) \nonumber\\
&= \frac{n N (N-1)}{2}
+ 2 \sum_{i<j} \sum_{a<b} s_i^a s_j^a s_i^b s_j^b .
\end{align}
Introducing the constant
\(
\Omega \equiv \exp\!\big(J_0^2 N(N-1)/4\big)
\)
to absorb the \(\mathcal{O}(n)\) term, we obtain the compact form
\begin{equation}
\label{eq:Zjnavg}
\mathbb{E}_J \big[ Z_J^{\,n} \big]
= \Omega^{\,n} \sum_{\{s^a\}}
\exp\!\Big(
J_0^2 \sum_{i<j} \sum_{a<b} s_i^a s_j^a s_i^b s_j^b
+ h \sum_{a=1}^{n} \sum_{i=1}^{N} s_i^a
\Big).
\end{equation}

\paragraph{Geometric interpretation.}
Equation~\eqref{eq:Zjnavg} defines a plaquette four-spin interaction on the graph product \(K_N \times K_n\) (complete graphs on \(N\) and \(n\) vertices, respectively). The construction is self-dual under the exchange \(N \leftrightarrow n\).
\begin{figure}
    \centering
    \includegraphics[width=0.2\linewidth]{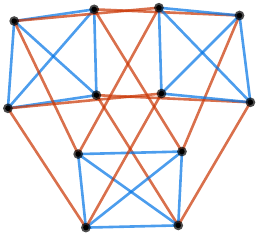}
    \caption{Example: $K_4 \times K_3$ graph.}
    \label{fig:placeholder}
\end{figure}

\subsection{Small-\texorpdfstring{$J_0$}{J0} expansion}\label{app:smallJ}

Assuming that the noise strength \(J_0\) is weak, we expand the replicated partition function in powers of \(J_0\):
\begin{align}
\mathbb{E}_J Z_J^{\,n}
&= \Omega^{n} \sum_{\{s\}}
e^{\,h \sum_{i,a} s_i^a}
\Bigg[
1 + J_0^2
\sum_{i<j,a<b}
s_i^a s_j^a s_i^b s_j^b
+ \mathcal{O}(J_0^4)
\Bigg].
\end{align}
Because the Boltzmann factor factorizes over spins,
\[
e^{h \sum_{i,a}s_i^a} = \prod_{i,a} e^{h s_i^a},
\]
the spin summations can be carried out independently. For a single spin, we define
\begin{equation}
z := \sum_{s=\pm1} e^{hs} = 2\cosh h,
\qquad
m := \langle s \rangle = \frac{1}{z}\sum_{s=\pm1} s\, e^{hs} = \tanh h.
\end{equation}
The term
\(
\big\langle \sum_{i<j,a<b}s_i^a s_j^a s_i^b s_j^b \big\rangle
= \tfrac{1}{4} N(N-1)n(n-1) m^4
\)
counts the number of plaquettes on the product graph \(K_N \times K_n\), corresponding to choosing one edge from \(K_N\) and one from \(K_n\) independently (\cref{fig:placeholder}); each summand represents a single four-spin plaquette interaction. Hence
\begin{equation}
\mathbb{E}_J Z_J^{\,n}
= \Omega^{n} (2\cosh h)^{nN}
\Big[
1 + \frac{J_0^2}{4}N(N-1)n(n-1)m^4 + \mathcal{O}(J_0^4)
\Big].
\label{eq:ZJn-expand}
\end{equation}

\paragraph{Replica limit.}
Taking the replica limit as in Eq.~\eqref{eq:replica-identity},
\begin{equation}
\mathbb{E}_J \log Z_J
= \log \Omega + N \log(2\cosh h)
- \frac{J_0^2}{4}N(N-1)\tanh^4 h
+ \mathcal{O}(J_0^4).
\label{eq:logZJ-final}
\end{equation}

\paragraph{Result for SAR.}
Substituting \eqref{eq:logZJ-final} into Eq.~\eqref{eq:log-sar-basic} yields
\begin{equation}
\log \mathrm{SAR}_{\rm geo}
= N\big(h - \log(2\cosh h)\big)
- \frac{J_0^2}{4}N(N-1)\big(1 - \tanh^4 h\big)
+ \mathcal{O}(J_0^4),
\label{eq:logSAR-expand}
\end{equation}
which is \cref{eq:leading} of the main text (note \(h-\log 2\cosh h=-\log(1+e^{-2h})\)). We have checked both coefficients against exact enumeration (\cref{app:numerics}): with a common set of Gaussian random numbers, \(\mathbb{E}_J\log p_J[\mathbf 1]\) is a smooth function of \(J_0\) whose \(J_0^2\) coefficient agrees with \eqref{eq:logSAR-expand} to within the statistical precision of the extraction (\(10\)--\(15\%\) for \(N=6\)--\(10\), \(h=1\)--\(3\)). The \(\mathcal{O}(J_0^4)\) coefficient, which involves five topological classes of plaquette pairs on \(K_N\times K_n\), was quoted in an earlier version of this work; the numerical extraction shows that the expansion is asymptotic only for \(J_0^2 N\ll1\), well before the crossover, and we therefore do not rely on it and evaluate the model exactly instead.

For \(h \gg 1\), using \(\tanh h \approx 1 - 2e^{-2h}\), the leading contribution in $e^{-2h}$ reads
\begin{equation}\label{eq:logSAR-largeh}
\log \mathrm{SAR}_{\rm geo}
\simeq
- e^{-2h}\, N
\Big[\,1 + 2J_0^2(N-1) + \mathcal{O}(J_0^4)\,\Big].
\end{equation}
The \(J_0^2 N(N-1)\) contribution is the super-linear (correlation) term; its sign is fixed and positive in $-\log\mathrm{SAR}$, so that disorder always accelerates the decay. To this order the model predicts the quadratic law \(\exp(-aN-bN^2)\) of \cref{sec:comparison}. Exponentiating the relative correction,
\begin{equation}\label{eq:emp-resum}
\log \mathrm{SAR} \simeq -e^{-2h} N\big[1 + 2J_0^2(N-1)\big] \;\to\; -e^{-2h}\,N\,e^{2J_0^2(N-1)},
\end{equation}
reproduces the form of the empirical law \cref{emp_sar} with $\beta_0 = e^{-2h}$ and $\alpha = e^{2J_0^2}$. We stress that \cref{eq:emp-resum} is a phenomenological closed form rather than a resummation of the perturbative series; the exact curves of \cref{app:numerics} show that it places the crossover too early and makes it too sharp. Its usefulness is limited to motivating the two-parameter form of \cref{emp_sar} and to interpreting \(\beta_0\) and \(\alpha\) qualitatively.

\subsection{Direct finite-\texorpdfstring{$N$}{N} evaluation}\label{app:numerics}

\begin{figure}[t]
    \centering
    \includegraphics[width=0.98\linewidth]{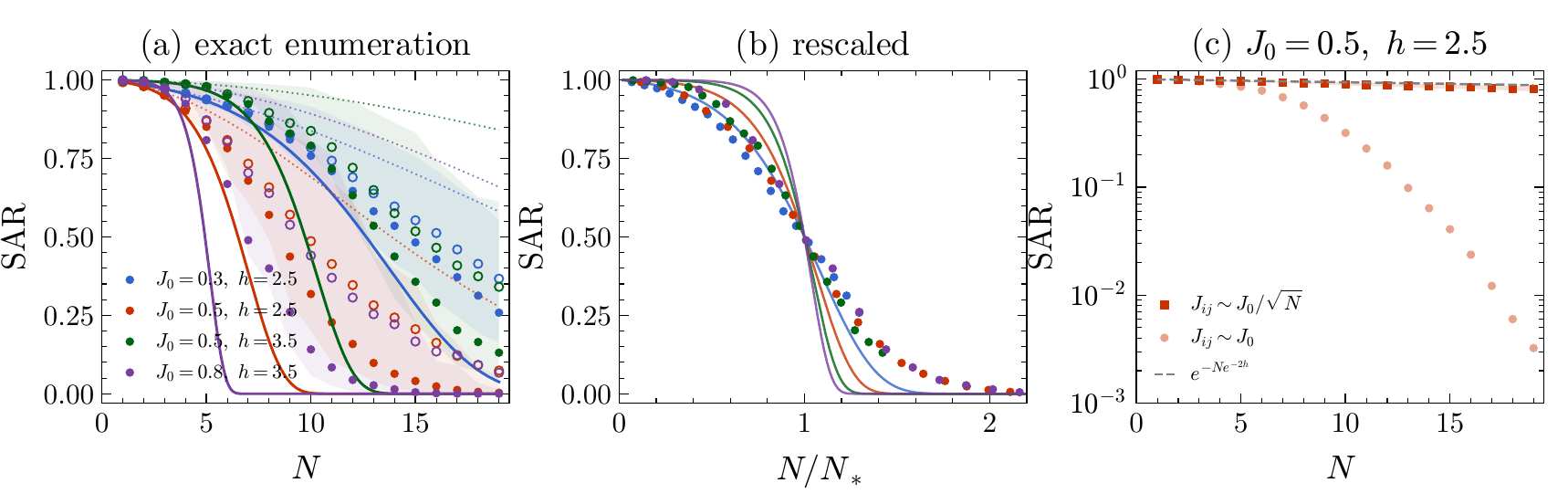}
    \caption{Direct finite-$N$ evaluation of the random-coupling model. (a) $\mathrm{SAR}(N)$ from exhaustive enumeration ($N\le19$, $200$ disorder realisations) for four parameter sets; filled symbols: geometric average $\exp\mathbb E_J\log p_J[\mathbf 1]$, open symbols: arithmetic average $\mathbb E_J p_J[\mathbf 1]$, bands: inter-quartile range over realisations. Solid lines: the exponentiated closed form \cref{eq:emp-resum} with the same $(J_0,h)$ and no free parameter; dotted: the leading-order quadratic law \cref{eq:logSAR-largeh}. (b) The exact geometric curves against $N/N_*$, together with the rescaled closed form for the corresponding $s=2J_0^2N_*$. (c) Contrast for $(J_0,h)=(0.5,2.5)$ between the $N$-independent coupling variance used in this work (circles) and the conventional SK normalisation $J_{ij}\sim J_0/\sqrt N$ (squares, log scale): with the latter, $-\log\mathrm{SAR}$ is linear in $N$ and no cliff develops.}
    \label{fig:sk-numerics}
\end{figure}

To assess the model beyond perturbation theory we evaluate $p_J[s=\mathbf 1]=e^{-E_J[\mathbf 1]}/Z_J$ exactly, by exhaustive enumeration of all $2^N$ configurations for $N\le 19$, and average over $M=200$ realisations of the couplings. To obtain smooth curves the same Gaussian random numbers are used for every $N$ (the couplings for size $N$ are the leading $N\times N$ block of one $19\times19$ realisation), so that neighbouring lengths share the disorder. Both the geometric average $\mathrm{SAR}_{\rm geo}$ of \cref{eq:logSAR-master} and the arithmetic average $\mathrm{SAR}_{\rm ari}=\mathbb E_J\,p_J[\mathbf 1]$ are recorded. \cref{fig:sk-numerics}(a) shows the results for $(J_0,h)=(0.3,2.5)$, $(J_0,h)=(0.5,2.5)$, $(J_0,h)=(0.5,3.5)$, $(J_0,h)=(0.8,3.5)$, chosen so that the crossover falls inside the enumerable range; the shaded bands are the inter-quartile ranges of $p_J[\mathbf 1]$ over realisations.

The exact curves display the qualitative phenomenology of the data: a plateau near $\mathrm{SAR}=1$ followed by a crossover whose scale is set by $J_0$ and $h$ and is an order of magnitude shorter than the independent-error scale $N_*^{(\alpha=1)}=\ln 2\,e^{2h}$ (\cref{tab:sk}). The crossover is well described by the two-parameter form \cref{emp_sar} with effective parameters, but the effective accumulation factor $\alpha_{\rm eff}$ is markedly smaller than the $e^{2J_0^2}$ suggested by exponentiating the leading-order correction: the closed form \cref{eq:emp-resum} places the crossover $15$--$30\%$ too early and makes it far too sharp, while the un-exponentiated quadratic law \cref{eq:logSAR-largeh} is accurate only for $N$ well below $N_*$ and places the crossover too late. The rescaled exact curves (\cref{fig:sk-numerics}(b)) have widths $w=0.55$--$0.66$ in units of $N_*$ (for the three sets whose crossover is completed within $N\le19$), i.e.\ the model produces a crossover at the sharp end of the range $w=0.6$--$1.6$ found for the LLM data; the broader crossovers of most model--task pairs are not reproduced by the model with a single $(J_0,h)$. The arithmetic average exceeds the geometric one through the crossover (by $0.07$--$0.2$ at $N\simeq N_*$, growing with $J_0$) and its crossover scale is $10$--$35\%$ larger, but the two show the same qualitative behaviour. Finally, \cref{fig:sk-numerics}(c) shows that with the conventional SK normalisation $J_{ij}\sim J_0/\sqrt N$ the disorder merely renormalises the exponential rate ($-\log\mathrm{SAR}/N\simeq0.0109$, compared with $\log(1+e^{-2h})=0.0067$ without couplings) and no cliff develops; the super-linear accumulation is a consequence of the $N$-independent coupling variance assumed in \cref{sec:model}, as stated there.

\begin{table}[t]
\centering
\caption{Crossover scales of the exactly enumerated model. $N_*^{\rm geo}$, $N_*^{\rm ari}$: from the geometric and arithmetic disorder averages; ``closed'': the exponentiated form \cref{eq:emp-resum}; ``quad'': the leading-order quadratic law; $\alpha=1$: independent errors, $\ln2\,e^{2h}$. $\alpha_{\rm eff}$ is the accumulation factor of a fit of \cref{emp_sar} to the exact geometric curve, to be compared with $e^{2J_0^2}$; $w$ is the crossover width of the exact curve in units of $N_*$.}
\label{tab:sk}
\begin{tabular}{lrrrrrrrr}
\toprule
$(J_0,h)$ & $N_*^{\rm geo}$ & $N_*^{\rm ari}$ & closed & quad & $\alpha=1$ & $e^{2J_0^2}$ & $\alpha_{\rm eff}$ & $w$ \\
\midrule
$(0.3,2.5)$ & 14.7 & 16.2 & 12.6 & 21.7 & 103 & 1.20 & 1.108 & -- \\
$(0.5,2.5)$ & 8.5 & 9.8 & 6.5 & 13.9 & 103 & 1.65 & 1.113 & 0.66 \\
$(0.5,3.5)$ & 13.4 & 15.3 & 9.7 & 38.5 & 760 & 1.65 & 1.155 & 0.55 \\
$(0.8,3.5)$ & 6.9 & 9.4 & 4.9 & 24.5 & 760 & 3.60 & 1.145 & 0.64 \\
\bottomrule
\end{tabular}
\end{table}

\section{Comparison of crossover laws}\label{app:comparison}

\cref{tab:comparison} lists, for every model--task curve, the quasi-AIC differences between the double-exponential law \cref{emp_sar} and (i) its $\alpha=1$ special case, (ii) the quadratic law $\exp(-aN-bN^2)$, (iii) the stretched exponential $\exp[-(N/N_0)^\gamma]$ and (iv) the logistic crossover $[1+e^{(N-N_0)/w}]^{-1}$. All laws are fitted by maximising the binomial log-likelihood $\sum_N[k_N\log p(N)+(n_N-k_N)\log(1-p(N))]$ of the instance counts; the quasi-AIC is $\mathrm{QAIC}=2P-2\log L/\hat c$, with $P$ the number of fitted parameters and $\hat c$ the design effect of the curve (\cref{sec:experiment}), and a positive $\Delta\mathrm{QAIC}$ means that the alternative fits worse than \cref{emp_sar}. Quasi-BIC gives the same ranking; the leave-one-length-out predictive likelihood agrees on the rejection of $\alpha=1$ on every curve and selects the same best two-parameter law on 13 of the 19 curves, the stretched and quadratic laws exchanging places on the remaining six. The last column gives the crossover scale $N_*$ of \cref{emp_sar} together with the range spanned by the four two-parameter laws. The $\alpha=1$ law is rejected on every curve ($\Delta\mathrm{QAIC}\ge35$); among the two-parameter laws the stretched exponential has the lowest QAIC on 14 curves, \cref{emp_sar} on 3 and the logistic on 2; the quadratic law is never the best but has a lower QAIC than \cref{emp_sar} on 14 of the 19 curves. The parameters of \cref{emp_sar}, their standard errors and the design effects for every curve are provided with the analysis code.

\begin{table}[t]
\centering
\caption{Model comparison for every model--task curve. $n$: number of instances; $\hat c$: batch design effect; $\Delta\mathrm{QAIC}$ of the $\alpha=1$ exponential, the quadratic, stretched-exponential and logistic laws relative to \cref{emp_sar} (positive: worse); $N_*$ of \cref{emp_sar} and its range over the four two-parameter laws.}
\label{tab:comparison}
\small
\resizebox{\linewidth}{!}{\input{model_comparison}}
\end{table}

\section{Proof of Theorem \ref{theorem1}}\label{app:proof}
\begin{proof}
Start with the condition
\begin{equation}\label{eq:NDC}
N\geq N_{\mathrm{DC}} \;=\; 1 + \frac{1}{\log\alpha}\left(\log\left(1-\frac{2\log\theta_{N,k}}{\beta_0}\right)+\frac{\log 2}{1-1/k}\right),
\end{equation}
with $\alpha>1$, $\beta_0>0$, $0< \theta_{N,k}\leq 1$, $k\geq 2$. Each term in $N_\mathrm{DC}$ is non-negative,
\begin{equation}
    \log\alpha > 0, \quad \log\left(1-\frac{2\log\theta_{N,k}}{\beta_0}\right)\geq 0,\quad \frac{\log 2}{1-1/k}>0,
\end{equation}
therefore Eq.~\eqref{eq:NDC} implies the following weaker conditions:
\begin{equation}\label{eq:ineq1}
    N> 1,
\end{equation}
\begin{equation}\label{eq:ineq2}
    N > 1+\frac{1}{\log\alpha}\log\left(1-\frac{2\log\theta_{N,k}}{\beta_0}\right)>1+\frac{1}{\log\alpha}\log\left(-\frac{2\log\theta_{N,k}}{\beta_0}\right),
\end{equation}
\begin{equation}\label{eq:ineq3}
    N> \frac{1}{\log\alpha}\frac{\log 2}{1-1/k},
\end{equation}
which are nevertheless more directly useful.

Eq.~\eqref{eq:ineq3} implies
\begin{equation}\label{eq:1/2bound}
\begin{split}
    &N(1-1/k)\log\alpha > \log 2 \\
    \Rightarrow\;&\alpha^{N(1-1/k)} > 2 \\\Rightarrow\;&\alpha^{-N(1-1/k)} < 1/2 \\\Rightarrow\;&(1-\alpha^{N/k-N})> 1/2.
\end{split}
\end{equation}
Eq.~\eqref{eq:ineq2} implies
\begin{equation}\label{eq:expbound}
    (N-1)\log\alpha>\log\left(-\frac{2\log\theta_{N,k}}{\beta_0}\right)
    \Rightarrow\;\frac{1}{2}\beta_0\alpha^{N-1}>-\log\theta_{N,k}.
\end{equation}
Putting Eqs.~\eqref{eq:ineq1}, \eqref{eq:1/2bound}, \eqref{eq:expbound} together, we have
\begin{equation}
\begin{split}
    \beta_0 N(\alpha^{N-1}-\alpha^{N/k-1})&=\beta_0 \alpha^{N-1} N (1-\alpha^{N/k-N})\\
    &>\beta_0 \alpha^{N-1} \times 1\times \frac{1}{2}\\
    &>-\log\theta_{N,k},
\end{split}
\end{equation}
therefore the logarithmic gain $\Delta(N,k)=\log\theta_{N,k}+\beta_0 N(\alpha^{N-1}-\alpha^{N/k-1})>0$ is positive.
\end{proof}

\end{document}

%% file: model_comparison.tex
\begin{tabular}{llrrrrrrr}
\toprule
task & model & $n$ & $\hat c$ & $\Delta\mathrm{QAIC}_{\alpha=1}$ & $\Delta\mathrm{QAIC}_{\rm quad}$ & $\Delta\mathrm{QAIC}_{\rm str}$ & $\Delta\mathrm{QAIC}_{\rm log}$ & $N_*$ (range) \\
\midrule
letter-4 & \texttt{gemini-2.5-pro} & 4900 & 1.9 & +260 & -20 & -33 & -21 & 12 (11--12) \\
letter-4 & \texttt{gemini-2.5-flash} & 4580 & 1.0 & +252 & -88 & -94 & -110 & 7 (7--8) \\
letter-9 & \texttt{gemini-2.5-pro} & 4900 & 1.7 & +292 & -33 & -34 & -9 & 17 (17--17) \\
letter-9 & \texttt{gemini-2.5-flash} & 4890 & 1.6 & +207 & -35 & -36 & -15 & 10 (9--10) \\
letter-13 & \texttt{gemini-2.5-pro} & 4900 & 1.5 & +257 & -36 & -37 & +7 & 21 (20--21) \\
letter-13 & \texttt{gemini-2.5-flash} & 4880 & 1.3 & +171 & -83 & -88 & -72 & 10 (10--11) \\
letter-26 & \texttt{gemini-2.5-pro} & 4900 & 1.2 & +425 & -44 & -46 & -19 & 19 (18--19) \\
letter-26 & \texttt{gemini-2.5-flash} & 4900 & 1.8 & +121 & -52 & -60 & -44 & 10 (10--11) \\
letter-13 & \texttt{gpt-5} & 3010 & 4.2 & +64 & +1 & -1 & +4 & 167 (148--167) \\
letter-13 & \texttt{grok-4} & 8960 & 2.5 & +270 & +9 & +11 & +13 & 55 (53--55) \\
letter-13 & \texttt{claude-4-sonnet} & 4900 & 1.5 & +293 & -79 & -80 & -49 & 16 (15--16) \\
addition & \texttt{gemini-2.5-pro} & 24830 & 1.2 & +1599 & +37 & +4 & +105 & 344 (332--344) \\
addition & \texttt{gemini-2.5-flash} & 8970 & 1.5 & +391 & -26 & -30 & +36 & 79 (75--80) \\
addition & \texttt{grok-4} & 2860 & 2.1 & +137 & +22 & +30 & +9 & 73 (68--73) \\
addition & \texttt{claude-4-sonnet} & 22400 & 1.0 & +982 & -48 & -52 & +163 & 244 (232--247) \\
Pauli (strict) & \texttt{gemini-2.5-pro} & 7970 & 4.8 & +147 & -3 & -4 & -8 & 89 (84--89) \\
Pauli (relaxed) & \texttt{gemini-2.5-pro} & 8000 & 5.2 & +182 & +2 & -21 & -4 & 104 (97--104) \\
Pauli (strict) & \texttt{gemini-2.5-flash} & 7950 & 4.9 & +35 & -10 & -17 & -13 & 32 (32--37) \\
Pauli (relaxed) & \texttt{gemini-2.5-flash} & 8000 & 5.9 & +56 & -22 & -28 & -16 & 41 (41--45) \\
\bottomrule
\end{tabular}

%% file: ref.bib
@article{wan2025cognitive,
  title={A cognitive writing perspective for constrained long-form text generation},
  author={Wan, Kaiyang and Mu, Honglin and Hao, Rui and Luo, Haoran and Gu, Tianle and Chen, Xiuying},
  journal={arXiv preprint arXiv:2502.12568},
  year={2025}
}

@inproceedings{kim2024llm,
  title={An LLM Compiler for Parallel Function Calling},
  author={Kim, Sehoon and Moon, Suhong and Tabrizi, Ryan and Lee, Nicholas and Mahoney, Michael W and Keutzer, Kurt and Gholami, Amir},
  booktitle={International Conference on Machine Learning},
  pages={24370--24391},
  year={2024},
  organization={PMLR}
}

@inproceedings{qian2024chatdev,
  title={ChatDev: Communicative Agents for Software Development},
  author={Qian, Chen and Liu, Wei and Liu, Hongzhang and Chen, Nuo and Dang, Yufan and Li, Jiahao and Yang, Cheng and Chen, Weize and Su, Yusheng and Cong, Xin and others},
  booktitle={Proceedings of the 62nd Annual Meeting of the Association for Computational Linguistics (Volume 1: Long Papers)},
  pages={15174--15186},
  year={2024}
}

@article{li2024survey,
  title={A survey on LLM-based multi-agent systems: workflow, infrastructure, and challenges},
  author={Li, Xinyi and Wang, Sai and Zeng, Siqi and Wu, Yu and Yang, Yi},
  journal={Vicinagearth},
  volume={1},
  number={1},
  pages={9},
  year={2024},
  publisher={Springer}
}

@inproceedings{xie2025fire,
  title={FIRE: Fact-checking with Iterative Retrieval and Verification},
  author={Xie, Zhuohan and Xing, Rui and Wang, Yuxia and Geng, Jiahui and Iqbal, Hasan and Sahnan, Dhruv and Gurevych, Iryna and Nakov, Preslav},
  booktitle={Findings of the Association for Computational Linguistics: NAACL 2025},
  pages={2901--2914},
  year={2025}
}

@inproceedings{press2023measuring,
  title={Measuring and Narrowing the Compositionality Gap in Language Models},
  author={Press, Ofir and Zhang, Muru and Min, Sewon and Schmidt, Ludwig and Smith, Noah A and Lewis, Mike},
  booktitle={Findings of the Association for Computational Linguistics: EMNLP 2023},
  pages={5687--5711},
  year={2023}
}

@inproceedings{wang2023plan,
  title={Plan-and-Solve Prompting: Improving Zero-Shot Chain-of-Thought Reasoning by Large Language Models},
  author={Wang, Lei and Xu, Wanyu and Lan, Yihuai and Hu, Zhiqiang and Lan, Yunshi and Lee, Roy Ka-Wei and Lim, Ee-Peng},
  booktitle={Proceedings of the 61st Annual Meeting of the Association for Computational Linguistics (Volume 1: Long Papers)},
  pages={2609--2634},
  year={2023}
}

@inproceedings{yao2023react,
  title={React: Synergizing reasoning and acting in language models},
  author={Yao, Shunyu and Zhao, Jeffrey and Yu, Dian and Du, Nan and Shafran, Izhak and Narasimhan, Karthik and Cao, Yuan},
  booktitle={International Conference on Learning Representations (ICLR)},
  year={2023}
}

@inproceedings{havrilla2024glore,
  title={GLoRe: When, Where, and How to Improve LLM Reasoning via Global and Local Refinements},
  author={Havrilla, Alexander and Raparthy, Sharath Chandra and Nalmpantis, Christoforos and Dwivedi-Yu, Jane and Zhuravinskyi, Maksym and Hambro, Eric and Raileanu, Roberta},
  booktitle={International Conference on Machine Learning},
  pages={17719--17733},
  year={2024},
  organization={PMLR}
}

@article{kojima2022large,
  title={Large language models are zero-shot reasoners},
  author={Kojima, Takeshi and Gu, Shixiang Shane and Reid, Machel and Matsuo, Yutaka and Iwasawa, Yusuke},
  journal={Advances in neural information processing systems},
  volume={35},
  pages={22199--22213},
  year={2022}
}

@article{wei2022chain,
  title={Chain-of-thought prompting elicits reasoning in large language models},
  author={Wei, Jason and Wang, Xuezhi and Schuurmans, Dale and Bosma, Maarten and Xia, Fei and Chi, Ed and Le, Quoc V and Zhou, Denny and others},
  journal={Advances in neural information processing systems},
  volume={35},
  pages={24824--24837},
  year={2022}
}

@article{madaan2023self,
  title={Self-refine: Iterative refinement with self-feedback},
  author={Madaan, Aman and Tandon, Niket and Gupta, Prakhar and Hallinan, Skyler and Gao, Luyu and Wiegreffe, Sarah and Alon, Uri and Dziri, Nouha and Prabhumoye, Shrimai and Yang, Yiming and others},
  journal={Advances in Neural Information Processing Systems},
  volume={36},
  pages={46534--46594},
  year={2023}
}

@article{wan2025fano,
  title={A Fano-Style Accuracy Upper Bound for LLM Single-Pass Reasoning in Multi-Hop QA},
  author={Wan, Kaiyang and Gao, Lang and Mu, Honglin and Nakov, Preslav and Wang, Yuxia and Chen, Xiuying},
  journal={arXiv preprint arXiv:2509.21199},
  year={2025}
}

@article{astolfi2024consistency,
  title={Consistency-diversity-realism Pareto fronts of conditional image generative models},
  author={Astolfi, Pietro and Careil, Marlene and Hall, Melissa and Ma{\~n}as, Oscar and Muckley, Matthew and Verbeek, Jakob and Soriano, Adriana Romero and Drozdzal, Michal},
  journal={arXiv preprint arXiv:2406.10429},
  year={2024}
}

@inproceedings{papineni-etal-2002-bleu,
    title = "{B}leu: a Method for Automatic Evaluation of Machine Translation",
    author = "Papineni, Kishore  and
      Roukos, Salim  and
      Ward, Todd  and
      Zhu, Wei-Jing",
    editor = "Isabelle, Pierre  and
      Charniak, Eugene  and
      Lin, Dekang",
    booktitle = "Proceedings of the 40th Annual Meeting of the Association for Computational Linguistics",
    month = jul,
    year = "2002",
    address = "Philadelphia, Pennsylvania, USA",
    publisher = "Association for Computational Linguistics",
    url = "https://aclanthology.org/P02-1040/",
    doi = "10.3115/1073083.1073135",
    pages = "311--318"
}

@article{liang2022holistic,
  title={Holistic evaluation of language models},
  author={Liang, Percy and Bommasani, Rishi and Lee, Tony and Tsipras, Dimitris and Soylu, Dilara and Yasunaga, Michihiro and Zhang, Yian and Narayanan, Deepak and Wu, Yuhuai and Kumar, Ananya and others},
  journal={arXiv preprint arXiv:2211.09110},
  year={2022}
}

@inproceedings{lin-2004-rouge,
    title = "{ROUGE}: A Package for Automatic Evaluation of Summaries",
    author = "Lin, Chin-Yew",
    booktitle = "Text Summarization Branches Out",
    month = jul,
    year = "2004",
    address = "Barcelona, Spain",
    publisher = "Association for Computational Linguistics",
    url = "https://aclanthology.org/W04-1013/",
    pages = "74--81"
}

@article{lewkowycz2022solving,
  title={Solving quantitative reasoning problems with language models},
  author={Lewkowycz, Aitor and Andreassen, Anders and Dohan, David and Dyer, Ethan and Michalewski, Henryk and Ramasesh, Vinay and Slone, Ambrose and Anil, Cem and Schlag, Imanol and Gutman-Solo, Theo and others},
  journal={Advances in neural information processing systems},
  volume={35},
  pages={3843--3857},
  year={2022}
}

@article{hendrycks2021measuring,
  title={Measuring mathematical problem solving with the math dataset},
  author={Hendrycks, Dan and Burns, Collin and Kadavath, Saurav and Arora, Akul and Basart, Steven and Tang, Eric and Song, Dawn and Steinhardt, Jacob},
  journal={arXiv preprint arXiv:2103.03874},
  year={2021}
}

@article{wang2022self,
  title={Self-consistency improves chain of thought reasoning in language models},
  author={Wang, Xuezhi and Wei, Jason and Schuurmans, Dale and Le, Quoc and Chi, Ed and Narang, Sharan and Chowdhery, Aakanksha and Zhou, Denny},
  journal={arXiv preprint arXiv:2203.11171},
  year={2022}
}

@article{liu2023lost,
  title={Lost in the middle: How language models use long contexts},
  author={Liu, Nelson F and Lin, Kevin and Hewitt, John and Paranjape, Ashwin and Bevilacqua, Michele and Petroni, Fabio and Liang, Percy},
  journal={arXiv preprint arXiv:2307.03172},
  year={2023}
}

@article{turpin2023language,
  title={Language models don't always say what they think: Unfaithful explanations in chain-of-thought prompting},
  author={Turpin, Miles and Michael, Julian and Perez, Ethan and Bowman, Samuel},
  journal={Advances in Neural Information Processing Systems},
  volume={36},
  pages={74952--74965},
  year={2023}
}

@inproceedings{lake2018generalization,
  title={Generalization without systematicity: On the compositional skills of sequence-to-sequence recurrent networks},
  author={Lake, Brenden and Baroni, Marco},
  booktitle={International conference on machine learning},
  pages={2873--2882},
  year={2018},
  organization={PMLR}
}

@article{cobbe2021training,
  title={Training verifiers to solve math word problems},
  author={Cobbe, Karl and Kosaraju, Vineet and Bavarian, Mohammad and Chen, Mark and Jun, Heewoo and Kaiser, Lukasz and Plappert, Matthias and Tworek, Jerry and Hilton, Jacob and Nakano, Reiichiro and others},
  journal={arXiv preprint arXiv:2110.14168},
  year={2021}
}

@article{mccoy2023embers,
  title={Embers of autoregression: Understanding large language models through the problem they are trained to solve},
  author={McCoy, R Thomas and Yao, Shunyu and Friedman, Dan and Hardy, Matthew and Griffiths, Thomas L},
  journal={arXiv preprint arXiv:2309.13638},
  year={2023}
}
